\documentclass[12pt]{article}

\usepackage{scicite}
\usepackage{times}
\usepackage{bbm}
\usepackage{xcolor}
\usepackage{graphicx}
\usepackage{booktabs}
\usepackage{mathtools}
\usepackage{amssymb}
\usepackage{hyperref}
\usepackage{subcaption}
\usepackage{float}
\usepackage{comment}
\usepackage[toc]{appendix}

\definecolor{darkblue}{rgb}{0,0.08,0.45}
\hypersetup{%
  pdfborder=0 0 0,
  colorlinks=true,
  citecolor=darkblue,
  urlcolor=darkblue
}

\newcommand{\app}{\raise.17ex\hbox{$\scriptstyle\sim$}}

\newenvironment{sciabstract}{%
\begin{quote} \bf}
{\end{quote}}

\title{Learning Humanoid Locomotion \\ over Challenging Terrain}
\author{%
Ilija Radosavovic, Sarthak Kamat, Trevor Darrell, Jitendra Malik\\[2mm]
\normalsize{University of California, Berkeley}\\
\normalsize{\href{https://humanoid-challenging-terrain.github.io/}{Project Page}}
}
\date{}

\begin{document} 
\baselineskip24pt
\maketitle 

\begin{sciabstract}
Humanoid robots can, in principle, use their legs to go almost anywhere. Developing controllers capable of traversing diverse terrains, however, remains a considerable challenge. Classical controllers are hard to generalize broadly while the learning-based methods have primarily focused on gentle terrains. Here, we present a learning-based approach for blind humanoid locomotion capable of traversing challenging natural and man-made terrain. Our method uses a transformer model to predict the next action based on the history of proprioceptive observations and actions. The model is first pre-trained on a dataset of flat-ground trajectories with sequence modeling, and then fine-tuned on uneven terrain using reinforcement learning. We evaluate our model on a real humanoid robot across a variety of terrains, including rough, deformable, and sloped surfaces. The model demonstrates robust performance, in-context adaptation, and emergent terrain representations. In real-world case studies, our humanoid robot successfully traversed over 4 miles of hiking trails in Berkeley and climbed some of the steepest streets in San Francisco.
\end{sciabstract}

\begin{figure*}
\centering
\includegraphics[width=\linewidth]{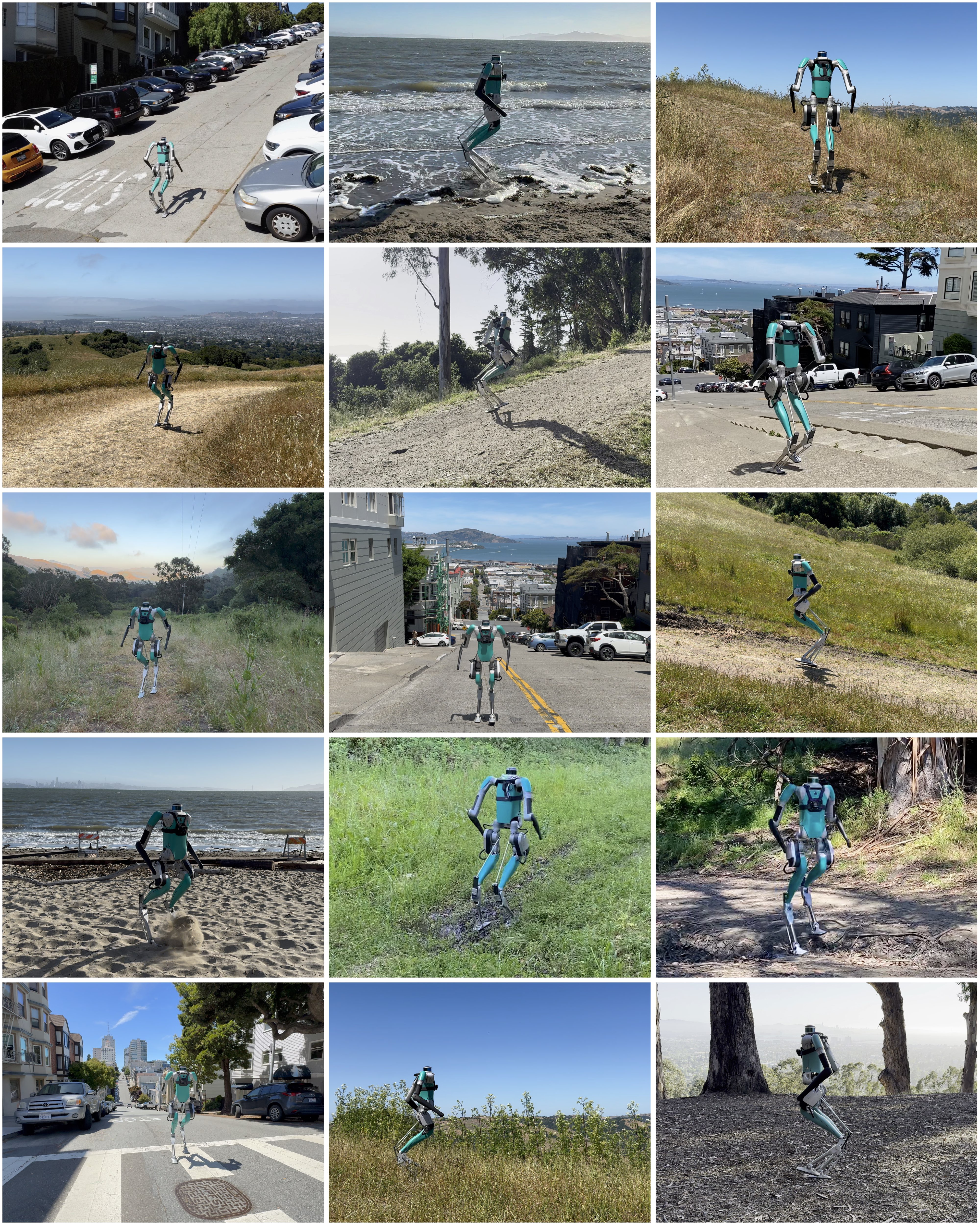}
\caption{\textbf{Humanoid locomotion over challenging terrain.} Our controller can successfully traverse a range of outdoor terrain, including steep, rough, rutted, rocky, wet, muddy, and sand.}
\label{fig:teaser}
\end{figure*}

\section*{Introduction}

Humanoid robots have the potential to operate in a variety of diverse environments on Earth, and beyond. A significant fraction of these involves walking over challenging terrains outdoors. These include natural environments like forests, deserts, and mountains, as well as man-made environments like city parks, cracked roads, and steep streets. Developing controllers for humanoid locomotion over challenging terrain, however, remains a significant challenge.

Legged locomotion has been studied extensively in the context of quadrupedal~\cite{raibert2008bigdog,zucker2010optimization,zico2011stanford,di2018dynamic,Lee2020,Kumar2021} and bipedal~\cite{zhao2012three,hereid20163d,hubicki2018walking,apgar2018fast,westervelt2018feedback,Gong2019,kim2020dynamic,Siekmann2021sim,Siekmann2021blind} robots. We now have quadrupedal robots that are able to traverse a variety of challenging terrains. Bipedal locomotion, for robots such as Cassie, is more challenging. And humanoid locomotion is even more challenging. The heavy legs, upper body with a lot of mass and inertia, and arms carrying payloads all contribute to the complexity of maintaining balance. These additional difficulties are further exacerbated when walking over challenging terrain.

Humanoid controllers have demonstrated reliable locomotion in structured environments using a variety of techniques~\cite{vukobratovic1968contribution,kajita2003biped,hirose2007honda,vukobratovic2004zero,collins2001three,tedrake2004stochastic,iida2004humanoid,collins2005efficient,stephens2010dynamic,johnson2015team,Kuindersma2016,koolen2016design,rodriguez2021deepwalk,Castillo2021}. Two popular paradigms are broadly based on the zero moment point principle~\cite{vukobratovic2004zero}, such as the Honda ASIMO, and model-based optimization~\cite{Kuindersma2016}, such as the Boston Dynamics Atlas. These methods are typically extended to uneven terrains by incorporating footstep planning with perception~\cite{chestnutt2003planning,deits2014footstep, griffin2019footstep}. However, they are often hard to generalize and adapt to new environments, or new variations of seen environments.

We have seen exciting results in learning-based real-world humanoid locomotion~\cite{RealHumanoid2023}. The paper proposed a transformer model that takes the past history of proprioceptive observations and actions as input, and predicts the next action. The model was trained with large-scale reinforcement learning in simulation and deployed it to real world zero-shot. The resulting controller can navigate various, albeit gentle, real-world terrains, including different surfaces, like concrete and grass, overground obstructions like debris and small steps, and slight slopes.

\newpage

This suggests a natural question: can we extend the same method to learning humanoid locomotion over challenging terrain? In principle, yes. We could increase the complexity of the training terrains in simulation, train the model for longer, and iterate on the sim-to-real transfer. In practice, however, extending this approach to challenging terrains is not straightforward and involves a set of mutually reinforcing challenges: an order of magnitude more training samples, more involved environment design, and more complex sim-to-real transfer.

Recently, we have seen an alternate approach to~\cite{RealHumanoid2023} that uses the same model architecture but proposes a different training method~\cite{radosavovic2024humanoid}. Instead of reinforcement learning, the model is trained with sequence modeling on a dataset of trajectories of walking on flat. Compared to reinforcement learning, sequence modeling is far more efficient and stable. However, the model can only be as good as the data it was trained on, which in this case means gentle terrains.

The aforementioned method is reminiscent of, and indeed inspired by, generative pre-training in natural language processing~\cite{Radford2018,radford2019language,Brown2020,ouyang2022training}. Modern pre-trained language models can be used zero-shot, prompted with few examples, or fine-tuned with additional data for alignment or to acquire new capabilities. Our core observation is that this general principle should apply to humanoid locomotion as well. Intuitively, if a robot can walk on flat ground, it should be more efficient to fine-tune it on challenging terrains, than it would be to train a robot that cannot walk at all.

Here, we propose a method for learning humanoid locomotion over challenging terrain. Our model is a transformer that predicts the next action from the history of proprioceptive observations and actions. Our method involves a two-step training procedure. First, we pre-train the model with sequence modeling of flat-ground trajectories. We leverage a dataset constructed using a prior policy, a model-based controller, and human sequences. Second, we fine-tune the model with reinforcement learning on uneven terrains. Pre-training enables the model to vicariously absorb the skills from prior data and provides a good starting point for learning new skills efficiently. We call our model Humanoid Transformer 2, or HT-2 for short.

\newpage

We evaluate our approach on a Digit humanoid robot through a series of real-world and simulation experiments. We find that our policies enable robust walking over a variety of different terrains, shown in Figure~\ref{fig:teaser}. These include very steep, rough, rutted, and rocky roads, as well as wet, muddy, and sand surfaces. Note that some of these terrains, like loose sand, have not been seen during training. We use a single neural network controller for all terrains.

To evaluate our approach in real-world scenarios, we performed two case studies. First, we took the robot on five different hikes in the Berkeley area, shown in Figure~\ref{fig:hikes}, and successfully completed over 4 miles of hiking. Second, we tested our robot on very steep streets in San Francisco, shown in Figure~\ref{fig:sfstreets}. With over 31\% grade, these are among some of the steepest streets in the city and the world. Our model successfully navigated all tested settings.

We perform a series of studies to understand the properties of our method. First, we look at the latent representations of the model as the robot is walking over different terrains. We find that the representations cluster based on the terrain (Figure~\ref{fig:representations}). These representations are emergent and have not been hand-designed or supervised during training. Second, we study the abilities of our model to adapt based on context. We observe kinematic adaptation to different terrain slope (Figure~\ref{fig:slowadapt}) and dynamic adaptation to terrain material (Figure~\ref{fig:fastadapt}). Note that the in-context adaptation behaviors are emergent and have not been pre-programmed.

We compare our approach to a state-of-the-art learning-based controller~\cite{RealHumanoid2023} and find that our model consistently outperforms the prior work (Figure~\ref{fig:comparisons}). To understand the key design choices, we perform a series of ablation studies. We find that pre-training leads to considerable sample efficiency gains compared to training from scratch (Figure~\ref{fig:ablations_1}), fine-tuning substantially improves performance (Figure~\ref{fig:ablations_2}), and both exhibit good gait patterns (Figure~\ref{fig:ablations_3}).

The presented results demonstrate that a general learning method enables humanoid locomotion over challenging terrain. We hope that the presented methodology will help accelerate deployment of humanoid robots in the full richness and complexity of the physical world.

\section*{Results}

We summarize the results in Figure~\ref{fig:teaser} and Movie 1. We train a single model to control a robot to walk over a variety of different settings, including steep, rough, rocky, and soft terrains.

\subsection*{Challenging terrain}

Our goal is to develop a learning-based approach that can control a real humanoid robot to walk across a variety of challenging terrains. We briefly discuss the hardware platform used in the experiments and give a basic sense of our approach before presenting the results.

We perform all experiments on the Digit humanoid robot developed by Agility Robotics. Digit is approximately 1.6 m tall and weights about 45 kg. The robot has 36 degrees of freedom including the floating base. The arms have four joints each, all of which are actuated. Each leg has eleven measurable joints, six of which are actuated. The passive joints are connected via springs and a four-bar linkage mechanism, making them difficult to simulate accurately.

Our controller is a transformer neural network model. It takes the history of past proprioceptive observations and actions as input, and predicts the next action as output. The controller is blind and does not use vision. It supports omni-directional walking by following the desired velocity commands given as input. The model predicts desired joint positions and PD gains at 50 Hz, which are then converted to torques via a PD controller running at 2000 Hz.

We deploy our controller to the real robot and evaluate it in a number of outdoor scenarios, like shown in Figure~\ref{fig:teaser}. The experiments were performed over the course of about two weeks, starting around mid May of 2024. We consider a variety of natural environments including steep, rough, rutted, and rocky terrains. We further evaluate our controller on challenging man-made terrains, such as very steep city streets with over 31\% grade. Note that many of these terrains like vegetation, wood chip, mud, water, sand, and others were not seen during training. All of the presented results were achieved using a single neural network controller.

\begin{figure*}
\centering
\includegraphics[width=0.95\linewidth]{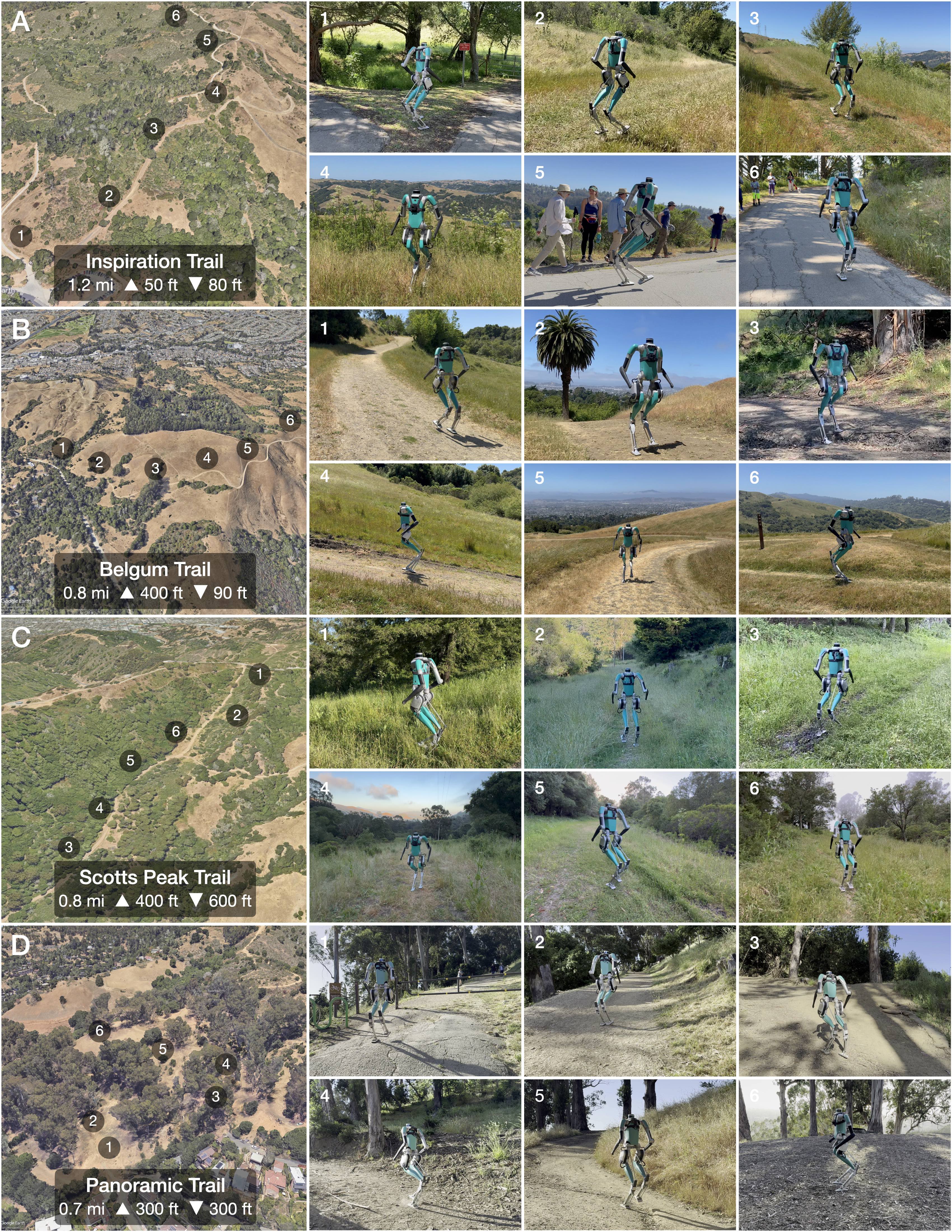}
\caption{\textbf{Berkeley hikes.} Our robot has successfully completed over 4 miles of real hiking trails in Berkeley. \textbf{(A-D)} We show basic statistics and examples from four different hikes.}
\label{fig:hikes}
\end{figure*}

\subsection*{Berkeley hikes}

To evaluate our approach in real-world scenarios, we perform two case studies. In the first study we take our robot to five different hikes in the Berkeley area, summarized in Figure~\ref{fig:hikes}. These are normal human hikes that are collectively about 4.3 miles long. Our policy was able to successfully complete all of the hikes. We describe each of the hikes in more detail next.

\emph{Lower Fire Trail.} The first hike we completed was the Lower Fire Trail. It is 0.83 miles long and has a 195 ft monotonically increasing elevation change. The terrain is a combination of rough and rocky ground, with tricky rutted areas from dried up water paths.

\emph{Panoramic Trail.} Our second hike was the Panoramic Trail, shown in Figure~\ref{fig:hikes}, D. The hike starts with a a long steady incline from the sidewalk over the dirt road, followed by a steeper incline over loose and rocky terrain, ending at a wood chip vista point at the top, and then going back to the starting point. A particularly challenging part was a very steep incline with loose and rocky ground, shown in image 3 from Figure~\ref{fig:hikes}, D.

\emph{Inspiration Trail.} We then completed the Inspiration Trail hike, shown in Figure~\ref{fig:hikes}, A. This was our longest hike at 1.2 miles. It contains gradual elevation changes over rough, rutted, and grass terrain, including rutted terrain covered in grass, and high grass, like shown in image 4.

\emph{Belgum Trail.} Our fourth hike was the Belgum Trail, shown in Figure~\ref{fig:hikes}, B. This hike consists of long steep slopes over rough and rocky terrain. The trail contains deep water marks, some of which were soft and muddy, like shown in image 3. The highest point of the trail is at a high altitude with strong wind causing nontrivial additional disturbances.

\emph{Scotts Peak Trail.} Our final hike was the Scotts Peak Trail, shown in Figure~\ref{fig:hikes}, C. This hike is 0.8 miles long and has large elevation changes. It starts with a long and steep descent over rough terrain with high grass (images 1--2), followed by a rocky and wet path through a forest (image 3), and loops back to the start (images 4--6). Parts of the trail are in deep shade and very wet, with sections covered in water and mud, like shown in image 3.

\begin{figure*}
\centering
\includegraphics[width=\linewidth]{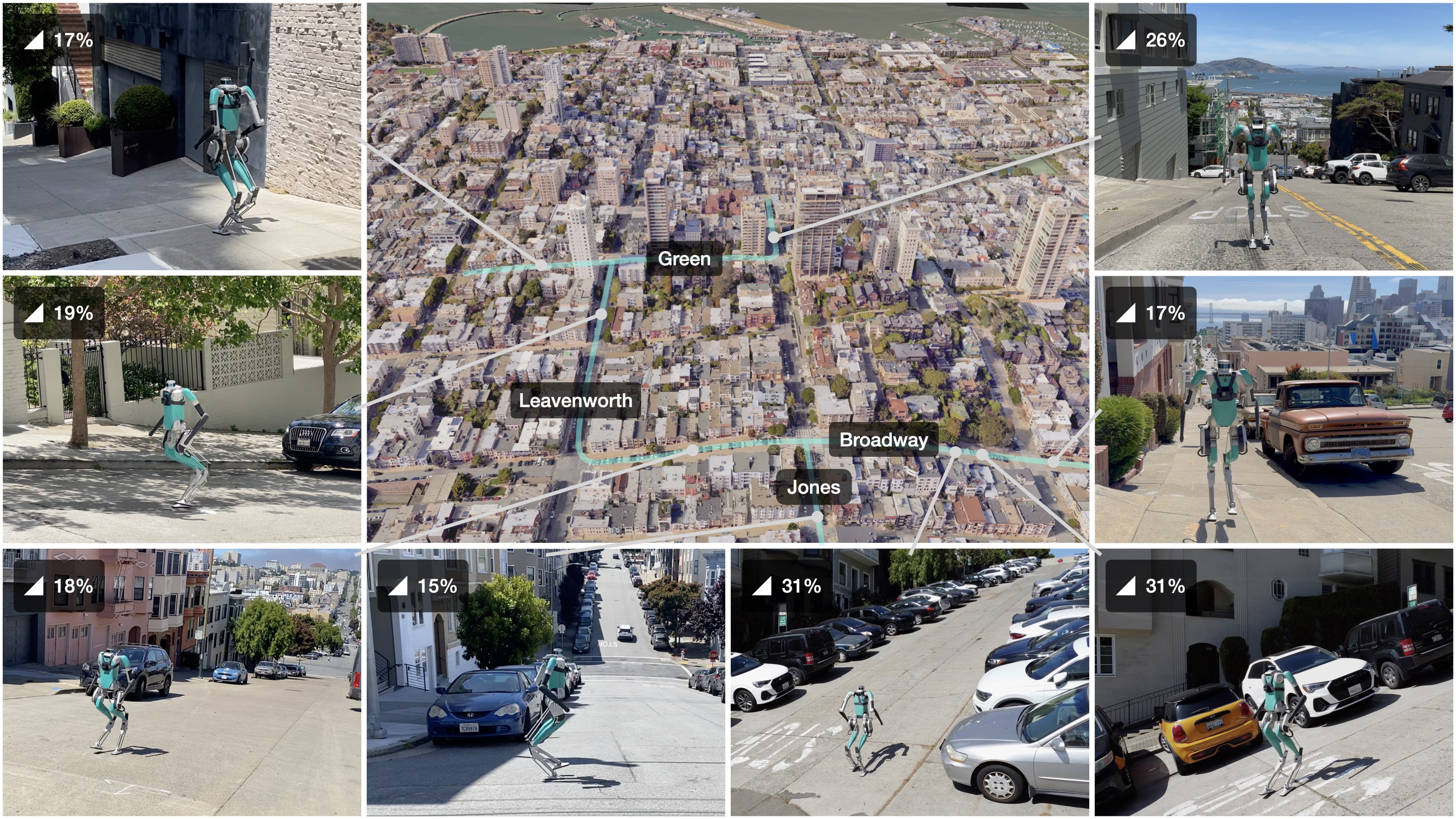}
\caption{\textbf{San Francisco streets.} Our robot can successfully navigate some of the steepest streets in San Francisco, which are also among some of the steepest streets in the world.}
\label{fig:sfstreets}
\end{figure*}

\subsection*{San Francisco streets}

In addition to walking over challenging terrains in natural environments, a strong locomotion controller should be able to handle challenging man-made terrains. In our second real-world study, we evaluate our controller in a challenging city environment.

In particular, we take our robot to the Russian Hill neighborhood of San Francisco, shown in Figure~\ref{fig:sfstreets}. This neighborhood is on a hill and is characterized by a number of very steep streets. Our controller was able to successfully navigate all of the tested streets. This included walking up and down steep roads, crossing uneven roads, and even turning on steep slopes.

In Figure~\ref{fig:sfstreets}, we show example images with corresponding slope grades. Note that these streets are very steep and hard even for healthy human adults. Moreover, the streets with slope grade of 31\% are among some of the steepest streets in San Francisco and in the world.

\subsection*{Terrain representations}

\begin{figure}
\centering
\includegraphics[width=\linewidth]{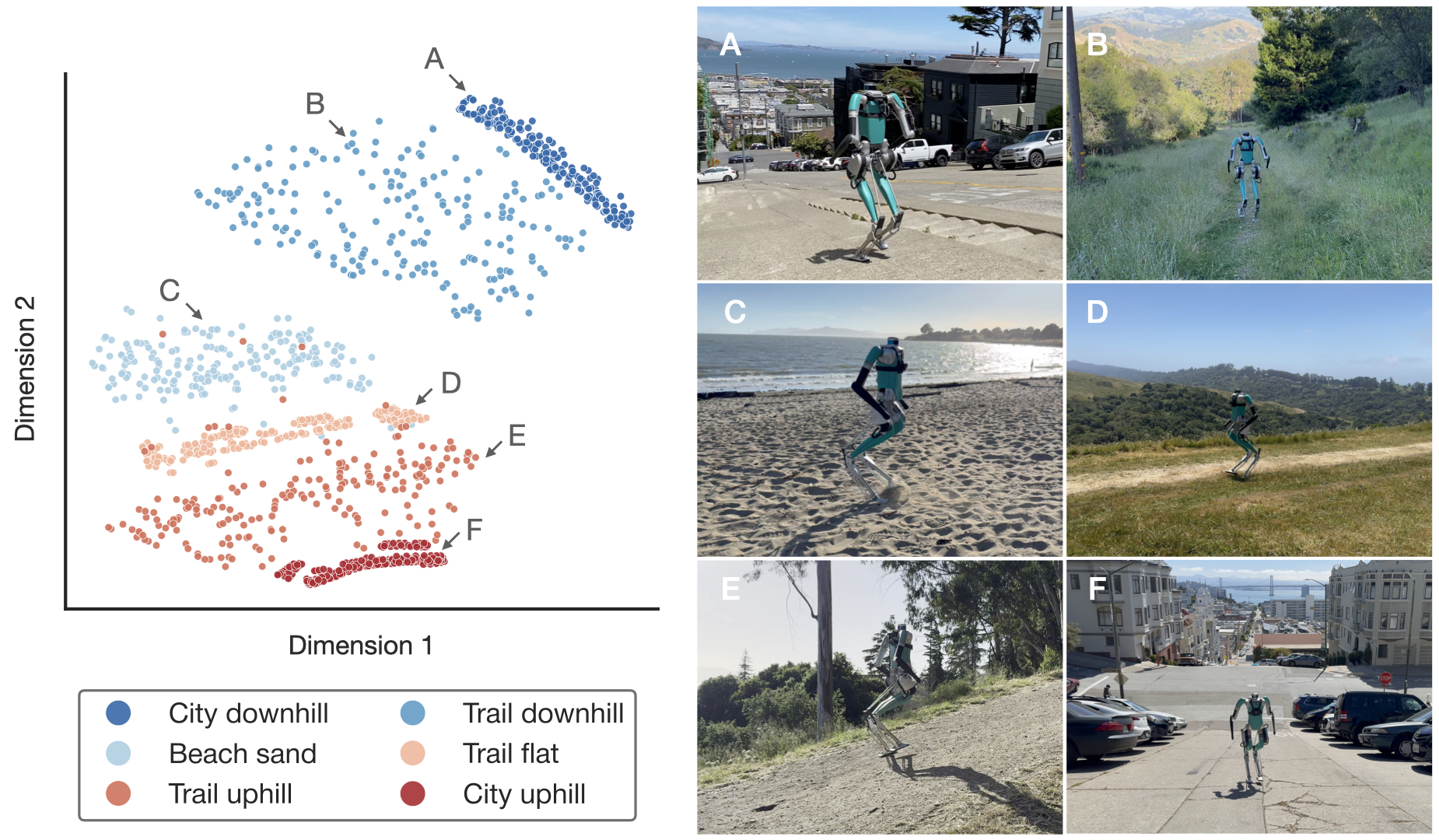}
\caption{\textbf{Terrain representations.} We observe that the representations of our model cluster based on the terrain the robot is walking on. These representations emerge as a byproduct of learning to walk. We did not design, supervise, or impose any constraints on the representations.}
\label{fig:representations}
\end{figure}

We have seen that our controller is able to traverse a variety of challenging terrains. To better understand the properties of our controller we perform a series of studies. We begin by analyzing the latent representations of our neural network model.

We record the latent representations from the last hidden layer of our transformer model while the robot is walking on six different terrains. We then project the latent representations to 2D using t-SNE~\cite{van2008visualizing} and plot them as a scatter plot in Figure~\ref{fig:representations}. We color code each point by the corresponding terrain type for visualization purposes only. We see that the points get grouped into six clear clusters corresponding to different terrain types. We note that these representations emerged as a byproduct of learning to walk and have not been hand-designed.

\subsection*{In-Context adaptation}

In order for the controller to traverse challenging terrain effectively it must be able to adapt its actions based on the environment. We find that our transformer model is able to do that through in-context adaptation. This behavior is emergent and has not been supervised during training. We discuss three different example scenario where this behavior manifests.

\subsubsection*{Kinematic adaptation to terrain slope}

\begin{figure}
\centering
\includegraphics[width=\linewidth]{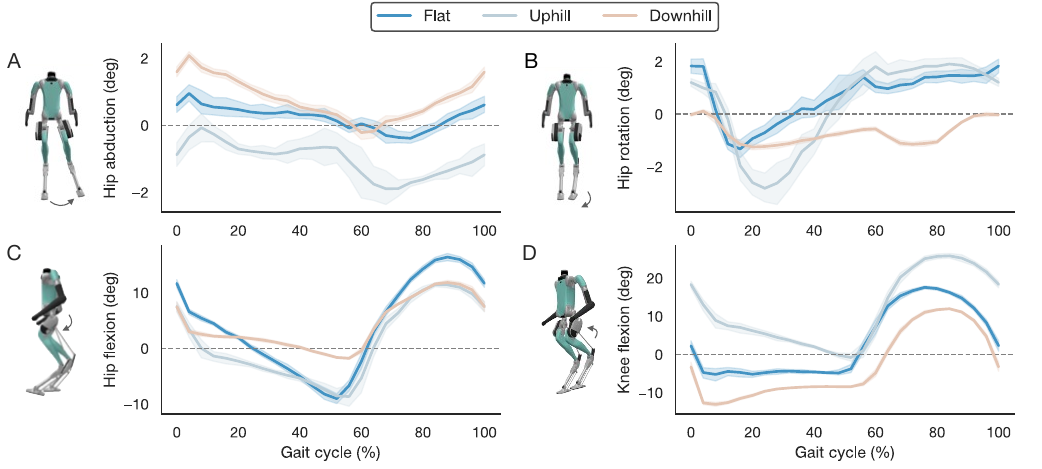}
\caption{\textbf{Kinematic adaptation.} We show the joint positions of the key hip and knee joints when walking over terrain with different slope. We see that our controller adapts its gait based on the terrain slope it is walking over. This has not been pre-programmed during training.}
\label{fig:slowadapt}
\end{figure}

We first study how the gait changes based on the terrain slope. In Figure~\ref{fig:slowadapt}, we show the joint positions of the core hip and knee joints over a gait cycle. Each curve corresponds to a different terrain slope, flat, uphill, and downhill. We observe that the controller is commanding different joint positions and consequently using different gaits based on the terrain slope. We note that this behavior is emergent and has not been pre-programmed during training. Namely, the model learned how to adjust its gait based on the terrain slope implicitly encoded in the context.

\begin{figure}
\centering
\includegraphics[width=\linewidth]{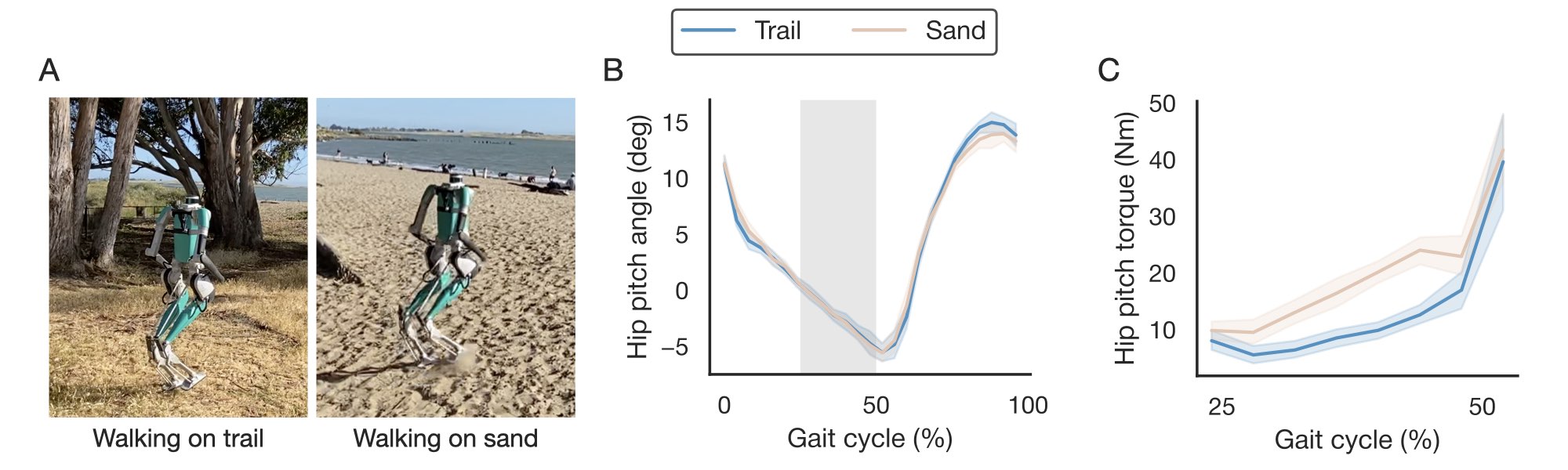}
\caption{\textbf{Dynamic adaptation.} \textbf{(A)} We study the behavior of our controller when walking on two different types of flat terrain, trail and sand. \textbf{(B)} We show the joint position of the core hip pitch joint over the gait cycle. We see that the controller achieves the same positions despite the considerable differences in the ground surface. \textbf{(C)} This is enabled by applying more torque when walking on sand to compensate for the dynamics. This shows that our controller can adapt at very short timescales in-context, based on the history of desired and achieved states.}
\label{fig:fastadapt}
\end{figure}

\subsubsection*{Dynamic adaptation to terrain material}

We study adaptation in the context of different terrain materials. We consider two settings, shown in Figure~\ref{fig:fastadapt}, A: walking on a hard surface and walking on sand. Human walking requires considerably more mechanical work and energy when walking on sand~\cite{lejeune1998mechanics}. In Figure~\ref{fig:fastadapt}, B, we see the achieved hip pitch positions over the gait cycle, which are the same in both settings. However, to achieve the same positions the controller exerts more torque on sand, as shown in Figure~\ref{fig:fastadapt}, C.  Recall that our controller is blind. However, it can still sense the terrain from the history of proprioceptive observations and actions, which encodes the discrepancy between the desired and achieved states and enables the model to adapt dynamically based on context.

\subsubsection*{Adaptation to external push disturbances}

In addition to adapting to terrain changes, a strong controller should also be able to adapt to external disturbances. In Movie 2, we apply strong external push while the robot is on uneven grass terrain. We see that the model quickly adapts its gait to prevent a fall and maintain balance.

\begin{figure}[t]
\centering
\includegraphics[width=1.0\linewidth]{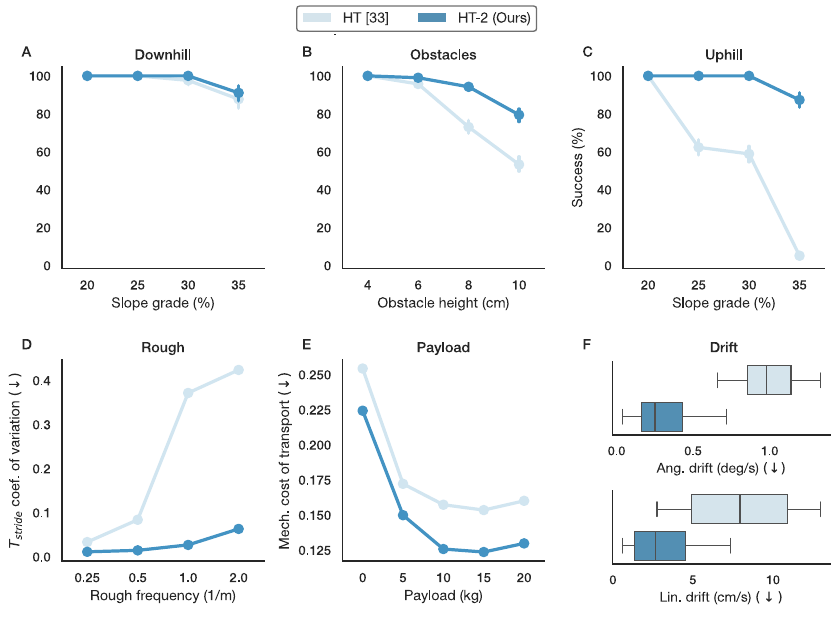}
\caption{\textbf{Comparison to baselines.} We compare the proposed HT-2 model to a state-of-the-art model HT~\cite{RealHumanoid2023}. Our controller has a higher success rate at traversing downhill \textbf{(A)}, obstacles \textbf{(B)}, and uphill \textbf{(C)} terrains. It is more efficient when walking over rough \textbf{(D)} and carrying different payloads uphill \textbf{(E)}, and is better at following velocity commands on flat ground \textbf{(F)}.}
\label{fig:comparisons}
\end{figure}

\subsection*{Comparisons}

In the previous sections, we have seen that the proposed controller can successfully walk over a range of challenging terrains. We now perform controlled quantitative comparisons to a state-of-the-art learning-based controller from prior work in simulation (Figure~\ref{fig:comparisons}). We perform the comparisons using the MuJoCo simulator~\cite{Todorov2012}, which supports equality constraints and enables us to simulate the complex structure of the Digit robot reasonably well. We repeat each experimental trial 256 times and report the mean and standard deviation for each experiment.

We compare our HT-2 model trained using the proposed method, to the Humanoid Transformer (HT) model from prior work~\cite{RealHumanoid2023}. For fair comparisons, we use the same transformer architecture hyper-parameters for both models. The models differ only in the way that they were trained. In the case of HT, we use the model trained by the authors. The model was trained with large-scale reinforcement learning in simulation for over a billion steps of walking over smooth planes, rough planes, and gentle slope terrains. In the case of HT-2, we trained the model via the proposed two-step approach. The model was first pre-trained with sequence modeling of flat-ground walking trajectories and then fine-tuned with reinforcement learning on uneven terrains, including hills, steep slopes, rough planes, and discrete obstacles.

We first compare the success rate of the models in traversing different terrain. We set up an environment with two flat sections separated by a terrain section in between. The robot starts on a flat section and is commanded to walk over the terrain section to the other flat section. A trial is successful if the robot reaches the goal in the allocated time duration. We consider three different terrain types with a range of difficulties for each. We report the results in Figure~\ref{fig:comparisons}, A--C. We see that the HT and HT-2 models are comparable downhill (A). The gap in performance increases over discrete obstacles (B) and is the largest uphill (C). Note that the performance of HT decays rapidly with increased difficulty while the HT-2 degrades gracefully.

Increased gait variability can be predictive of human falls and is often used in clinical studies, particularly with the elderly~\cite{hausdorff2001gait}. One way to quantify gait variability is to look at the stride time variability and measure the coefficient of variation. We adopt this metric to asses the performance of robot locomotion over rough terrain. The basic idea is that a controller that is better at handling rough terrain will stumble less and have a smaller coefficient of variation. We report the results in Figure~\ref{fig:comparisons}, D. We see that the HT-2 model has a considerably smaller stride time variability. Moreover, the HT-2 performance decreases gracefully with the increase in terrain difficulty. In contrast, HT performance degrades rapidly with increased difficulty.

In the next experiment we asses the performance of the controllers when walking up hill with a payload. Namely, we increase the mass of the robot by up to 20 kg and command the robot to walk uphill. We then measure the mechanical cost of transport. The controller that has a more efficient gait should have a lower mechanical cost of transport. In Figure~\ref{fig:comparisons}, E, we report the result. We see that HT-2 consistently outperforms HT across all payloads.

A good controller should be able to track desired velocity commands accurately. One way to quantify this is to measure amount of drift while following commands. In Figure~\ref{fig:comparisons}, F, we report linear drift in the ground XY plane and angular drift around the vertical Z axis. We see that the HT-2 model has a considerably lower drift making it more controllable.

\section*{Discussion}

The presented results show that a single neural network can learn to control a humanoid robot to walk over a variety of challenging terrains. These include rigid terrains seen during training as well as terrains such as mud, water, sand, grass, and many others, which were not seen during training. Our results were achieved using a general learning method that enables rapid acquisition of skills from existing datasets and efficient learning of new skills through interaction. We believe that the proposed methodology can have broad applications.

An important limitation and opportunity for future work is incorporating vision. Namely, our current controller is blind and relies solely on proprioceptive observations. While it is interesting that a blind controller can traverse the terrains shown here at all, incorporating vision has the potential to bring two major benefits. First, when walking over terrains with discrete irregularities, such a curb, a blind controller must either get lucky to step over them or ``sense'' them by bumping into them. Having access to vision as a distant sense could enable the model to anticipate the terrain changes and choose its steps accordingly. Second, it would make it feasible to navigate terrains for which vision is necessary, like stairs and stepping stones. A clear opportunity for future work is to use the presented methodology as a starting point for developing a locomotion controller that uses both visual and proprioceptive observations.

More broadly, the presented methodology can serve as a solid foundation for expanding capabilities of humanoid robots beyond locomotion. Our model architecture and data-driven training strategy make few domain-specific assumptions beyond the problem formulation. An exciting opportunity for future work is to use our methodology as a starting point for developing a unified humanoid model that can perform both locomotion and manipulation.

Overall, the presented results demonstrate that a general learning method enables humanoid locomotion over a variety of challenging terrains. We hope that the presented methodology will accelerate deployment of humanoid robots in unstructured real-world environments.

\section*{Materials and Methods}\label{sec:method}

We now present the methodology used to train the neural network controller that achieved the results from the previous sections. We describe the problem formulation, model architecture, our pre-training and fine-tuning procedure, and present ablation studies of key design choices.

\subsection*{Problem formulation}

We formulate the problem of blind humanoid locomotion as follows. The objective of our humanoid controller is to output motor torques to maintain balance while following desired velocity commands. The observation space consists of previous motor actions, joint positions, joint velocities, linear velocity of the base, angular velocity of the base, and velocity commands in the form of the desired linear velocity in the horizontal plane and the angular velocity around the up axis (Table~\ref{tab:obs_state}). Note that the observations include the positions and velocities of all joints, both actuated and passive. The action space is parameterized by a proportional-derivative (PD) controller, and consists of position, stiffness, and damping values for each motor (Table~\ref{tab:act_state}).

We perform all experiments on the Digit humanoid robot developed by Agility Robotics. Digit is approximately 1.6 m tall and weights about 45 kg. The robot has 36 degrees of freedom including the floating base. The arms have four joints each, all of which are actuated. Each leg has eleven measurable joints, six of which are actuated. The passive joints are connected via springs and a four-bar linkage mechanism, making them difficult to simulate accurately.

\subsection*{Model architecture}

We represent the controller as a neural network. Specifically, we use the Transformer~\cite{Vaswani2017} architecture, which has shown excellent results in a number of domains~\cite{Radford2018,radford2019language,Brown2020}, including humanoid locomotion~\cite{RealHumanoid2023,radosavovic2024humanoid}. This architecture has several appealing properties, two of which are particularly well-suited to the proposed methodology: \emph{(i) Scalable pre-training:} it allows for scalable pre-training on large datasets with general learning objectives; \emph{(ii) In-context learning:} it enables the model to learn to adapt its behavior at test time based on context.

Our model has few domain-specific biases beyond the choice of the inputs and the outputs. We provide the history of the previous $k$ proprioceptive observations and actions $o_{t-k}, a_{t-k}, ..., o_{t}$ as input. And the model predicts the next action $a_{t}$ as output. We encode the input observations and actions into a sequence of $k$ tokens. Specifically, for each timestep $t$, we first concatenate the observations and actions into a single vector. We then encode the concatenated vector into a vector of the same dimension as the hidden size of the transformer. We use a small multi-layer perceptron (MLP) as the encoder. The weights of the MLP encoder are shared across time. To encode the temporal order and the position of each token within the input sequence, we add the sinusoidal position embeddings to the input tokens. We then pass the sequence of $k$ tokens into the Transformer model. Our architecture follows a fairly standard Transformer design and consists of a sequence of $L$ layers. Each layer contains of a multi-head self-attention module and an MLP module. We encode temporal dependencies among the tokens by using causal attention. In particular, we use a causal mask in the multi-head self-attention modules to ensure that each token only attends to itself and previous tokens. Once the tokens are processed by the transformer, we apply an output projection head. The output projection head is a small MLP that projects an embedding from the hidden size of the transformer back to the original input dimension. Note that the predictions correspond to the concatenated inputs, containing both the observations and the actions, for each timestep. We apply the output projection head to all of the timesteps during pre-training, and only at the last timestep during fine-tuning. Similarly, the loss is applied to the full output predictions during pre-training, and only to the action subset during fine-tuning. We discuss pre-training and fine-tuning in the following sections.

Unless noted otherwise, the transformer model used in the experiments has a context window of 16 timesteps and four transformer blocks. Each block has a hidden dimension of 192, uses a multi-head self-attention module with four heads, and the MLP ratio of two. The input MLP encoder has hidden sizes of [512, 512] while the output MLP head has hidden sizes of [256, 128]. The transformer model has 1.4 million learnable parameters in total.

\subsection*{Pre-training with sequence modeling}

One option to train our model would be to perform reinforcement learning~\cite{RealHumanoid2023}. In other words, initialize the weights of the neural network randomly, from scratch, and use that as the starting point for trial and error learning. This process is hard to perform in the real world, at present, and can be performed in simulation instead. Namely, the model is trained on an ensemble of randomized environments and transferred to the real world. The overall process is delicate and time consuming, often requiring billions of samples in simulation, intricate reward design, and iterating the train-deploy loop to achieve successful sim-to-real transfer. These issues are further exacerbated with the complexity of the task, making our task of learning humanoid locomotion over challenging terrains particularly tricky. While some of these challenges can be partially alleviated through the use of techniques like distillation and privileged information~\cite{Openai2019,Lee2020,Kumar2021,RealHumanoid2023}, the fundamental challenges of learning through interaction from scratch still remain.

To overcome these challenges, our methodology uses a two step learning procedure. Our core observation is that even though the capabilities that we want to learn, in this case humanoid locomotion over challenging terrain, are not readily available in the form of data, we can still leverage the sources of available data to learn relevant skills. Recent work~\cite{radosavovic2024humanoid} has shown that we can train a humanoid model to walk on gentle terrains purely from offline data coming from prior neural network controllers, classical model-based controllers, and humans. We show that this method can serve as an effective pre-training procedure and enable efficient learning of new skills subsequently. Intuitively, a robot that already knows how to walk on flat ground should be better at learning how to walk up a slope, than a robot that cannot walk at all. Pre-training a humanoid model on existing trajectories is similar in spirit to pre-training a language model on the text from the Internet~\cite{Radford2018,radford2019language,Brown2020}. We describe the pre-training procedure next.

We begin by compiling a dataset $\mathcal{D}$ of humanoid trajectories. Each trajectory is a sequence of proprioceptive observations and actions over time: $\mathcal{T} = (o_1, a_1, o_2, a_2, ..., o_T, a_T)$. As a major source of trajectories we use a neural network policy from prior work~\cite{RealHumanoid2023}. We run the model and record trajectories in simulation. In each trajectory, the robot walks on flat ground for 10s and follows a randomly sampled velocity command. As an additional source of data we use trajectories that come from a model-based controller and humans. We use the model-based controller provided by the robot manufacturer in simulation. This controller is available via an application programming interface (API) and we do not get access to its internals. Like before, we collect 10s long trajectories of walking with randomly sampled commands. However, unlike in the case of a policy, we only get access to proprioceptive observations. Thus, each trajectory is a sequence of observations without the actions: $\mathcal{T} = (o_1, o_2, ..., o_T)$. In the case of human data, we use human walking trajectories from existing motion capture (MoCap) datasets~\cite{Plappert2016,AMASS:ICCV:2019} and human videos~\cite{carreira2017quo,andriluka2018posetrack}. We extract human poses from videos using computer vision techniques~\cite{rajasegaran2022tracking}, which can be seen as noisy MoCap. Each human walking trajectory is a sequence of human poses. Since the morphologies of the human and the robot are different, we retarget the human poses to the robot morphology via inverse kinematics (IK). This provides us with approximate proprioceptive observations for the joints and the base. We set the velocity commands for each trajectory approximately as well, using hindsight re-labeling. Like in the case of the model-based trajectories, we do not get access to actions for the human trajectories.

\begin{figure}[t]
\centering
\includegraphics[width=1.0\linewidth]{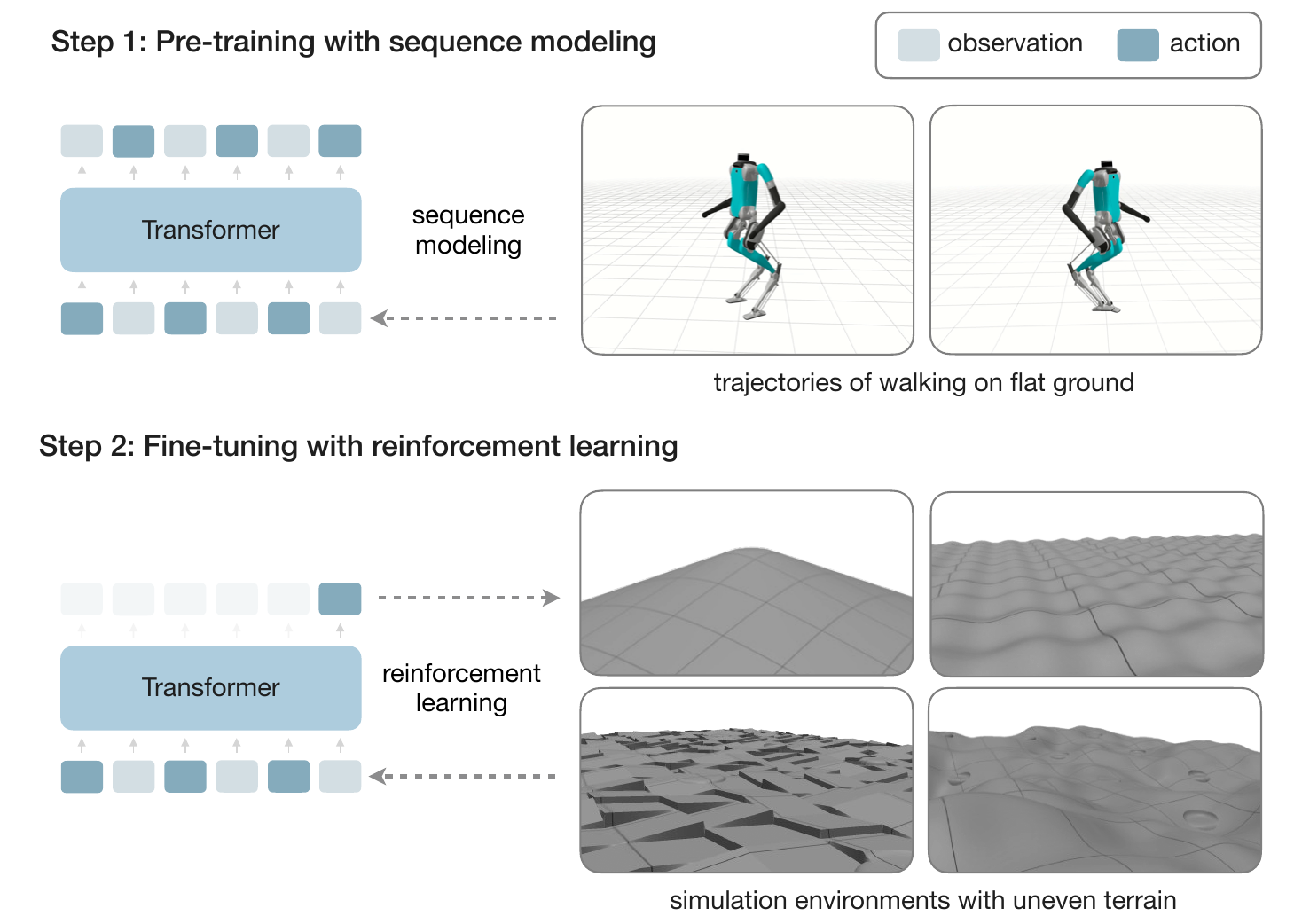}
\caption{\textbf{Method.} Our controller is a transformer model that predicts the next action from the history of past proprioceptive observations and actions. Our training procedure consists of a pre-training stage (top) and a fine-tuning stage (bottom). In the pre-training stage, we pre-train the model with sequence modeling on a dataset of trajectories of walking on flat ground. In the fine-tuning stage, we fine-tune the model with reinforcement learning on uneven terrains.}
\label{fig:method}
\end{figure}

We now describe our pre-training objective. Our goal is to train our transformer neural network architecture to model the joint trajectory distribution $p(\tau)$ autoregressively:
\begin{equation}
    p(\tau) = \prod_{k=1}^{K} p(a_k, o_k | a_{k-1}, o_{k-1}, ..., a_{1}, o_{1})
\end{equation}
To do that, we wish to minimize the negative log-likelihood over the dataset of trajectories:
\begin{equation}
    L = \sum_{\tau \in \mathcal{D}} - \log p(\tau)
\end{equation}
We assume a Gaussian distribution with constant variance and train our model to minimize the mean squared error between the predicted and the ground truth values. Our model takes in and predicts the normalized observation and action values in the continuous space. We found this approach to work well and do not perform quantization of the input or the output values.

In the exposition in the previous paragraph we assumed that each trajectory is a sequence of proprioceptive observations and actions. Recall, however, that a subset of our dataset containing trajectories that come from the model-based controller and humans does not have actions. To overcome this challenge, we replace the missing actions with learnable mask tokens $\texttt{[M]}$ to obtain $\mathcal{T} = (o_1, \texttt{[M]}, ..., o_T, \texttt{[M]})$. This enables us to tokenize and input all of the trajectories in a unified format. During training, we ignore the loss for the predictions that correspond to the masked parts of the inputs. The intuition is that sequences of observations, even without the actions, contain useful information about the world that the model can learn from. Moreover, by jointly training with the data that has actions, the model might learn how to inpaint the actions for the data without actions. And, consequently, model the joint distribution better.

We pre-train the model for 300 epochs on 4 A100 NVIDIA GPUs. We use a global mini-batch size of 4096. We employ a cosine learning rate schedule with the initial learning rate of 5e-4. The learning rate is warmed up linearly for 30 epochs starting from one tenth of the initial learning rate. We use the AdamW optimizer with weight decay of 0.01, $\beta_1$ of 0.9, and $\beta_2$ of 0.95. We summarize the pre-training hyperparameters in Table~\ref{tab:pre_hyperparams}.

\subsection*{Fine-tuning with reinforcement learning}

After pre-training on a dataset of trajectories, we obtain a model that is able to walk. However, since the model was trained on flat-walking trajectories its capabilities are limited mostly to gentle terrains. For example, the model cannot walk on steep slopes or terrains that are far from the pre-training dataset. Moreover, the model is not very robust to external disturbances or large changes in terrain properties. Nevertheless, as our experiments show, the pre-trained model can serve as an effective starting point for acquiring new capabilities through interaction.

To acquire new skills, we formulate a reinforcement learning (RL) problem in simulation. We assume access to a physics simulator with environment and robot states $s_t$, a transition function $S(s_{t+1} | s_t, a_t)$, and an observation function $O(o_t | s_t)$. We define a reward function $R(s_t, a_t, s_{t+1})$, consisting of terms that specify the locomotion task and encourage desirable properties, like smoothness and energy minimization. We provide the reward function terms and coefficients in the Supplement. We wish to learn a policy $\pi(a_t|o_t,a_{t-1}, ...,o_1)$ that maximizes the expected sum of discounted rewards through interactions with the environment.

We parameterize the policy as a diagonal multivariate Gaussian, with a mean represented by a neural network model and a learnable diagonal covariance matrix. We initialize the weights of the neural network with the pre-trained model. Pre-training can thus be viewed as a favorable initialization for reinforcement learning. At training time, we sample actions from the Gaussian distribution to enable the policy to explore. We use the mean actions predicted by the neural network at test time. Since our policy starts off from a pre-trained model that is already reasonable, we need to make sure that the initial exploration noise does not overwhelm the learning and instantly undo the effect of pre-training. Consequently, we randomly initialize the learnable covariance matrix to small values; smaller than those typically used for RL from scratch~\cite{RealHumanoid2023}. The hope is that this will enable the model to start exploring slowly, around the pre-trained model, and gradually learn to expand the exploration frontier to acquire new skills.

We use the proximal policy optimization (PPO) algorithm~\cite{Schulman2017} for fine-tuning the policy. PPO is a policy gradient method that in addition to training the policy, or the actor, involves training a critic network. Since the critic network is only used at training time and not required at deployment, we can take take advantage of the oracle state information available in simulation during training~\cite{Pinto2017,Openai2019,RealHumanoid2023}. In particular, the critic model receives the states rather than the observation as input. We summarize the information contained in the critic state in Table~\ref{tab:act_state}. The actor and the critic use the same network architecture, except for the first and the last layers which have different dimensions to account for different input and output dimensions. They do not share weights. The critic is initialized from scratch and is not pre-trained.

To acquire locomotion skills over uneven terrains, we fine-tune the model on a set of procedurally generated terrains: rough flats, smooth slopes, rough slopes, discrete obstacles, and hills. For each of the terrain types we consider variations in difficulty, such as the slope grades and obstacle heights. To make sure that the model retains a good flat walking gait, we include the flat terrain, resulting in six terrain types in total. We fine-tune the model on an ensemble of environments with different terrains jointly. Note that we fine-tune the model only on rigid terrains, without soft or deformable terrains. Nevertheless, as demonstrated in our results, models fine-tuned in this way can generalize to unseen real-world terrains during deployment. We summarize the terrain types, difficulty ranges, and proportions in Table~\ref{tab:terrain}.

We use a reward function that consists of multiple terms that can be divided into two groups. The first group specifies the locomotion task with velocity tracking and includes the terms for tracking linear velocity in the forward-backward and left-right directions, the angular velocity around the up direction, and an alive bonus. In principle, or in the real world, these terms should be sufficient to learn a good walking behavior. In practice, however, our simulation environments and compute are limited. We are unable to simulate physical phenomena, such as battery life and hardware damage, and the full diversity and richness of real-world terrains. To partially alleviate these challenges in the short term, we include additional reward terms. Specifically, we include energy, ground contact, base movement, foot airtime, foot symmetry, foot swing, and torso reward terms. The reward terms are summarized in the Supplement.

We fine-tune the model on a single A10 NVIDIA GPU. We train for 2000 iterations using 2048 parallel environments and 24 steps per environment, resulting in $\app$100M environment steps in total. We perform PPO updates over 5 epochs with a minibatch size of 12288. We set the PPO clipping parameter to 0.2, the weight decay to 0.01, and do not use the entropy regularization. We use the initial learning rate of 1e-5 and the cosine learning rate schedule for the actor. The learning rate is warmed up linearly for 100 iterations starting from 1e-8. The critic uses a constant learning rate of 5e-4. We use a discount factor $\gamma$ of 0.99 and generalised advantage estimation $\lambda$ of 0.95. We summarize the fine-tuning hyperparameters in Table~\ref{tab:ppo_hyperparams}.

\subsection*{Sim-to-real transfer}

We fine-tune our model with reinforcement learning in simulation and transfer it to the real world zero-shot. Our general strategy for sim-to-real transfer broadly follows~\cite{RealHumanoid2023}. We provide additional details on simulation, domain randomization, and real-world deployment below.

We perform experiments on a Digit humanoid robot. Digit uses four-bar linkages that introduce loops in the kinematic tree. We use the MuJoCo simulator~\cite{Todorov2012} which is able to simulate this using equality constraints. While MuJoCo is considerably slower than GPU-based simulators, like IsaacGym~\cite{Makoviychuk2021}, we can afford to use a slower simulator thanks to the sample efficiency of our method. In particular, our method only requires  $\app$100M samples for fine-tuning, which we are able to acquire in MuJoCo in about a day in our implementation. Specifically, we use a batched environment pool~\cite{weng2022envpool} to parallelize simulation on CPUs of a single machine. We model the knees via equality constraints. We treat the toes as passive for consistency with prior data. We leave potential improvements from controlling the toes for future work.

Each simulated environment consists of a procedurally generated terrain and a single robot tasked with following velocity commands. We fine-tune our model on an ensemble of different terrains. Specifically, we consider six different terrain types: flat, rough, smooth slope, rough slope, discrete obstacles, and hills. We pre-assign a probability to each terrain type and sample a terrain for each environment. We additionally sample terrain-specific parameters that induce variations within terrains of the same type, such as slope steepness or obstacle heights. We provide the summary of terrain types, parameters, and example images in Table~\ref{tab:terrain}.

The velocity commands consist of three values: forward linear velocity, lateral linear velocity, and turning angular velocity. We sample each value independently from a pre-specified range using a uniform distribution. Since moving in all three dimensions simultaneously is rarely required, and infeasible for many triplets, we zero out each command dimension with probability of one-half, for training efficiency. We summarize the commands in Table~\ref{tab:command}.

To enable the model to transfer from simulation to the real world, we fine-tune it on an ensemble of randomized environments. We randomize environment physics, such as gravity, friction, and damping ratios. To more accurately model real-world sensors, we apply noise and delays to sensor observations in simulation. In addition, we randomize the robot properties, including body mass and size, joint damping and stiffness, and actuation. We note that our goal is not to learn a robust controller that can control the robot in all of the environments in the same way. Rather, we train a transformer model that can adapt based on the context of the environment it is operating in. We summarize the randomization ranges in Table~\ref{tab:dr}.

After fine-tuning in simulation, we deploy our model to hardware zero-shot. The model runs on the CPU of the onboard Intel NUC computer. We obtain joint positions, velocities, and IMU information from the low-level robot API. We do not use any control software provided by the manufacturer. Our model produces actions at 50 Hz. The actions specify the joint position targets and the PD gains for the PD controller that runs at 2000 Hz, and outputs torques.

\subsection*{Ablation studies}

We analyze some of the key components of our methodology. We study pre-trained, fine-tuned, and scratch models through the lens of sample complexity, performance, and gait quality.

\begin{figure}
\centering
\includegraphics[width=1.0\linewidth]{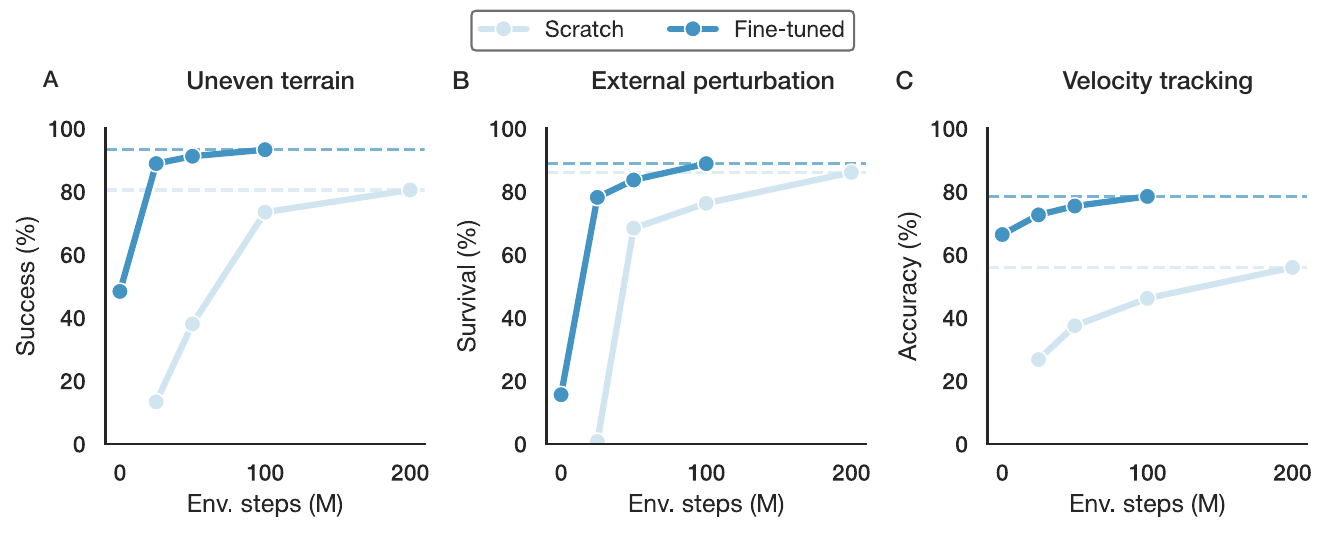}
\caption{\textbf{Comparison to scratch policy.} We compare a fine-tuned policy to a policy trained from scratch, as a function of the number of environment steps used for training. We include the scratch performance with up to $2\times$ more steps. We consider three different settings: uneven terrains, external perturbations, and velocity tracking. We observe that fine-tuning leads to a far better sample complexity and achieves considerably better absolute performance in all settings.}
\label{fig:ablations_1}
\end{figure}

\subsubsection*{Comparison to scratch policy}

We begin by studying the impact of fine-tuning a model compared to training a model from scratch. For fair comparisons, we optimize the hyper parameters of the scratch model separately. We optimize the fine-tuned and the scratch models for up to 100M and 200M environment steps, respectively. We consider three evaluation settings: uneven terrain, external perturbations, and velocity tracking. We report the performance as the function of training steps in Figure~\ref{fig:ablations_1}. We observe that fine-tuning a model leads to a far better sample complexity than training from scratch. Moreover, fine-tuning leads to considerably better absolute performance across all settings. This suggests that fine-tuning does not only accelerate but also improves learning.

\begin{figure}[t]
\centering
\includegraphics[width=\linewidth]{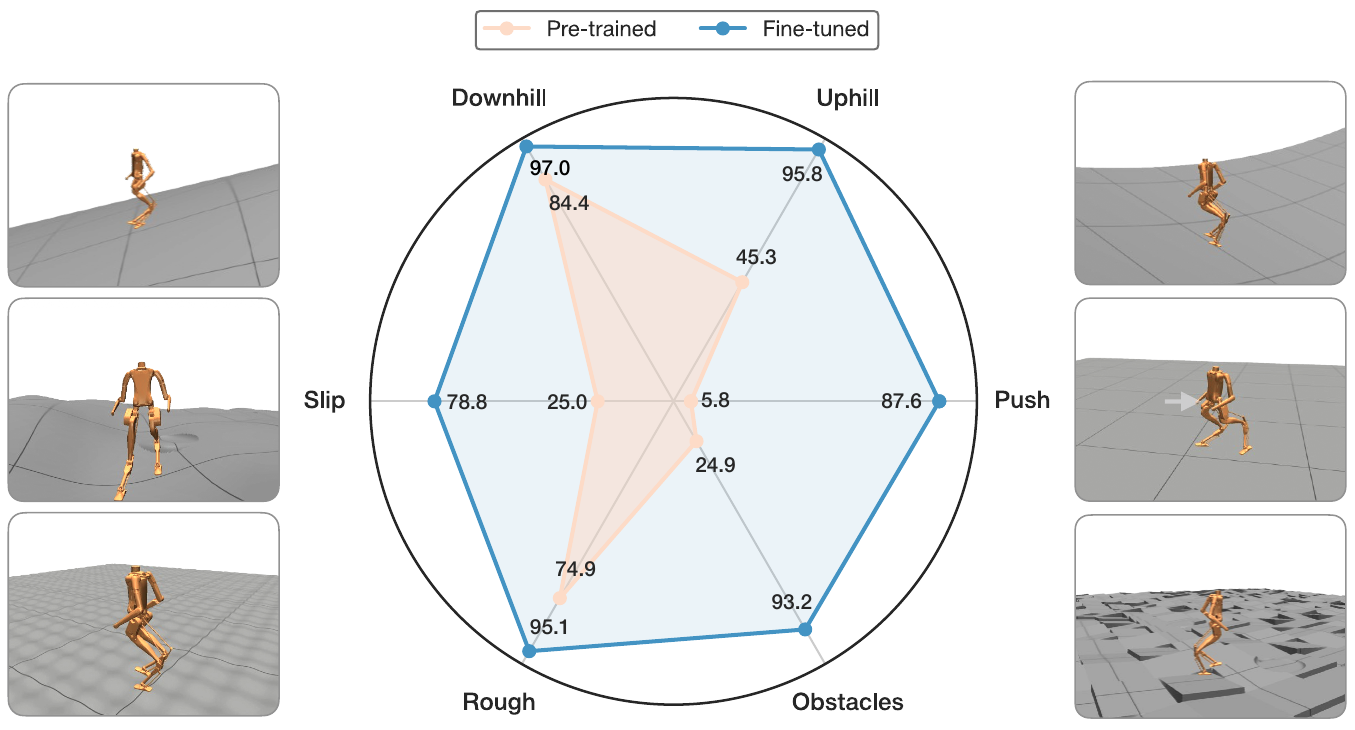}\vspace{4mm}
\caption{\textbf{Comparison to pre-trained policy.} We compare the performance of a pre-trained policy to a policy fine-tuned with reinforcement learning. We consider six different settings: downhill, uphill, push, obstacles, rough, and slip. We observe that the pre-trained policy performs reasonably well in easier settings, like downhill and rough. However, once we consider harder settings, we see that fine-tuning brings substantial improvements in performance; like on uneven terrains, such as uphill and obstacles, over slippery terrain, and with push disturbances.}
\label{fig:ablations_2}
\end{figure}

\subsubsection*{Comparison to pre-trained policy}

To understand the benefits fine-tuning with reinforcement learning brings over the pre-trained model, we compare a pre-trained policy to a fine-tuned policy. In Figure~\ref{fig:ablations_2}, we report the success rates in six different settings: downhill, uphill, push, obstacles, rough, and slip. First, we observe that the pre-trained policy performs reasonably well in easier settings, such as downhill and rough. Second, we observe that fine-tuning leads to substantial improvements in performance in harder settigns. These include uneven terrains, such as uphill and discrete obstacles, slippery terrain, and external push disturbances. These findings suggest that pre-training provides a reasonable starting point and that the RL fine-tuning leads to a substantially better policy.

\begin{figure}[t]
\centering
\includegraphics[width=1.0\linewidth]{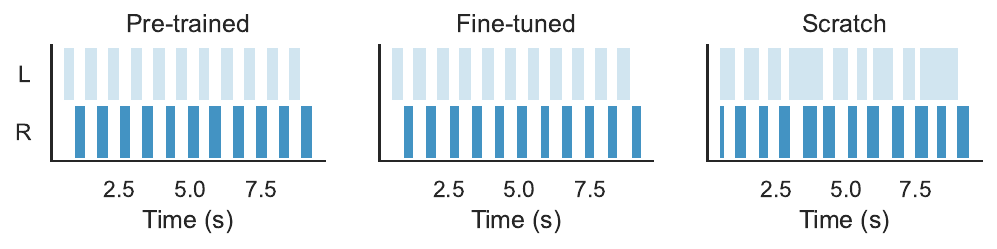}
\caption{\textbf{Comparison of gait patterns.} We compare the gait patterns for three models: pre-trained, fine-tuned, and scratch. We see that both the pre-trained and fine-tuned policy have good and similar gait patterns while the scratch policy has a considerably worse gait pattern.}
\label{fig:ablations_3}
\end{figure}

\subsubsection*{Comparison of gait patterns}

Gait quality has a big impact on locomotion performance. A good gait enables more capable, efficient, and visually pleasing locomotion. Moreover, it is less likely to be reliant on simulator imperfections and more likely to transfer to the real world. In Figure~\ref{fig:ablations_3}, we compare the gait patterns for three models: pre-trained, fine-tuned, and scratch. We see that both the pre-trained and the fine-tuned policy have good gaits that are fairly similar. The scratch policy, however, has a considerably worse gait. It is highly asymmetric and its left leg is being lifted up less and is dragging. For fair comparisons, both the fine-tuned and the scratch policy are trained on the same set of terrains and with the same set of rewards. These findings suggest that pre-training helps the model to acquire new capabilities during fine-tuning, while retaining a good gait.

\section*{Acknowledgments}

This work was supported in part by DARPA Machine Common Sense program, DARPA TIAMAT program (HR00112490425), ONR MURI program (N00014-21-1-2801), NVIDIA, and BAIR’s industrial alliance programs. We thank Daniel Yum, Rahul Meka, Sasha Ostojic, and Playground Global for providing the Digit robot used in the experiments; Bike Zhang and Torsten Darrell for help with the experiments; Koushil Sreenath for providing lab space.

\bibliography{scibib}
\bibliographystyle{IEEEtran}
\clearpage

\begin{appendices}
\setcounter{figure}{0}
\setcounter{table}{0}
\setcounter{footnote}{0}
\renewcommand{\thesection}{S\arabic{section}}
\renewcommand{\thefigure}{S\arabic{figure}}
\renewcommand{\thetable}{S\arabic{table}}
\renewcommand{\thesubfigure}{\Alph{subfigure}}
\section*{Supplementary Materials}

\subsection*{Reward function}\label{sec:reward}

Our reward function is a sum of the following terms:

\begin{itemize}

\item Linear velocity tracking reward ($r_{lv}$): This reward tracks the targets of forward and sideway walking velocity.
\begin{equation*}
    r_{lv} \coloneqq \mathrm{exp}(-||v_{xy} - v^*_{xy}||_2^2 / \sigma_{xy}),
\end{equation*}
where $v_{xy}$ and $v^*_{xy}$ represent the realized and commanded base linear velocity, respectively. We set $\sigma_{xy}$ to 0.2.

\item Angular velocity tracking reward ($r_{av}$): This reward tracks the target of turning velocity,
\begin{equation*}
    r_{av} \coloneqq \mathrm{exp}(-(\omega_z - \omega^*_z)^2 / \sigma_{\omega}),
\end{equation*}
where $\omega_z$ and $\omega^*_z$ represent the realized and commanded base angular velocity along the z-axis, respectively. We set $\sigma_{\omega}$ to 0.2.

\item Alive reward ($r_a$): rewards longer episode lengths.
\begin{equation*}
    r_a \coloneqq 1
\end{equation*}

\item Motor power reward ($r_{W}$): This reward penalizes output torques to reduce energy consumption and prevent hardware damage. 
\begin{equation*}
    r_{W} \coloneqq -5 \cdot 10^{-4} \cdot | \langle \tau \,, \dot{q} \rangle |,
\end{equation*}
where $\tau$ are the motor torques and $\dot{q}$ are the motor velocities.

\item Foot contact power reward ($r_{fcp}$): This reward penalizes overly high contact force on the robot's feet.
\begin{equation*}
    r_{fcp} \coloneqq 
    -0.002 \cdot \sum_{i \in \mathrm{foot}}{
      | \langle F(i) \,, v(i) \rangle |, }
\end{equation*}
where $F(i)$ is the foot contact force, and $v(i)$ is the foot velocity.

\item Base angular velocity reward ($r_{bav}$): This reward penalizes roll-pitch motions of the robot's base.
\begin{equation*}
    r_{bav} \coloneqq -0.25 \cdot ||\omega_{xy}||_2^2,
\end{equation*}
where $\omega_{xy}$ is the base's roll-pitch velocity.

\item Base linear velocity reward ($r_{blv}$): This reward penalizes vertical motion of the robot's base.
\begin{equation*}
    r_{blv} \coloneqq -1.5 \cdot ||v_z||_2^2,
\end{equation*}
where $v_z$ is the base linear velocity on the z-axis.

\item Base orientation reward ($r_{bo}$): This reward penalizes the base's orientation of the robot.
\begin{equation*}
    r_{bo} \coloneqq -0.5 \cdot ||g_{xy}||_2^2,
\end{equation*}
where $g_{xy}$ is the x-and-y component of the projected gravity vector.

\item Airtime reward ($r_{air}$): rewards the robot for lifting feet off the ground.
\begin{equation*}
    r_{air} \coloneqq \mathrm{min}([t_{air}, t_{last}, t_{max}])
\end{equation*}
where $t_{air}$ represents the cumulative air time of the swing.  $t_{last}$ is the  air time of the previous swing foot, and $t_{max} = 0.75$. Summing this reward over the duration of the step produces a quantity that is quadratic in the average air time.

\item Foot lead time symmetry ($r_{sym}$): rewards the robot for positioning the left and right foot symmetrically over time.
\begin{equation*}
    r_{sym} \coloneqq -0.25 \cdot \mathrm{max}([0, 1 - (t_{lead} / t_{lag})])
\end{equation*}
  We define a lead and lag foot as follows. If $\mathrm{foot}_{i}$ is currently in ahead of $\mathrm{foot}_{j}$ in the robot's base frame, $\mathrm{foot}_{i}$ is the lead foot, and $\mathrm{foot}_{j}$ is the lag foot. The cumulative  time the lead foot has been in front of the lag foot is defined to be $t_{lead}$. When the lead foot changes, we set $t_{lag}$ to the previous value of $t_{lead}$.

\item Footswing trajectory tracking ($r_{fs}$): This reward penalizes the deviation of the footswing trajectory from heuristic trajectories.
\begin{equation*}
    \begin{aligned}
        r_{fs} &\coloneqq 0.5 \cdot \sum_{i \in \mathrm{foot}}{\mathrm{exp}(- ||\mathrm{foot\_traj}_{i,xy} - \mathrm{foot\_traj}_{i,xy}^*||_2^2} / \sigma_{xy}) \\
        & - 50.0 \cdot \sum_{i \in \mathrm{foot}}{(\mathrm{foot\_traj}_{i,h} - \mathrm{foot\_traj}_{i,h}^*)^2},
    \end{aligned}
\end{equation*}
where $\mathrm{foot\_traj}_{xy}$ and $\mathrm{foot\_traj}_{h}$ are x-y and z-component of heuristic foot trajectories. The x-y component uses Raibert heuristics~\cite{Raibert1986} and the z-component uses von Mises distributions ($\kappa = 0.04$). We set $\sigma_{xy}$ to $0.2$.

\item Selected joint position penalty ($r_{jp}$): This reward penalizes deviations from a ``neutral" joint position for selected arm joints. 
\begin{equation*}
    r_{jp} \coloneqq - \sum_{j \in M}{\alpha_j(q(j) - q_0(j))^2},
\end{equation*}
where $q(j)$ and $q_0(j)$ represent the current and neutral joint positions of joint $j$, respectively. The joints penalized include shoulder's roll and yaw with a weight of $\alpha=5$, and shoulder pitch with a weight of $\alpha=0.25$.

\end{itemize}

\subsection*{Additional details}

We provide additional details on:
\begin{itemize}
  \item Observation and state spaces in Table~\ref{tab:obs_state}
  \item Action space in Table~\ref{tab:act_state}
  \item Pre-training hyperparameters in Table~\ref{tab:pre_hyperparams}
  \item Fine-tuning hyperparameters in Table~\ref{tab:ppo_hyperparams}
  \item Terrain types in Table~\ref{tab:terrain}
  \item Command ranges in Table~\ref{tab:command}  
 \item Domain randomization in Table~\ref{tab:dr}
\end{itemize}

\begin{table}
\begin{center}
\begin{tabular*}{0.66\linewidth}{@{}lrcc@{}}
\toprule
  Input & Dimensionality & Actor & Critic \\
\midrule
  Base Linear Velocity & 3 & \checkmark & \checkmark \\
  Base Angular Velocity & 3 & \checkmark& \checkmark \\
  Joint Positions & 26 & \checkmark & \checkmark \\
  Joint Velocities & 26 & \checkmark & \checkmark \\
  Projected Gravity & 3 & \checkmark & \checkmark \\
  Clock Input & 2 & \checkmark & \checkmark \\
  Commands & 3 & \checkmark & \checkmark \\
\midrule
  Denoised Projected Gravity & 3 & & \checkmark \\
  Command Drift & 3 & & \checkmark \\
  Foot Statistics  & 8 & & \checkmark \\
  Gait Guidance & 10 & & \checkmark \\ 
  Height Map & 121 & & \checkmark \\
  Relative Actions & 48 & & \checkmark \\
  Absolute Actions & 48 & & \checkmark \\
  Robot Parameters & 163 & & \checkmark \\
\bottomrule
\end{tabular*}
\caption{\textbf{Observation and state spaces.}  Both the pre-trained model and the actor take in the proprioceptive observations that are accessible on hardware during deployment. Since we do not require the critic at test time, we additionally provide the state information to the critic during fine-tuning. This state information is readily accessible in simulation and helps learning.}
\label{tab:obs_state}
\end{center}
\end{table}

\begin{table}
\begin{center}
\begin{tabular*}{0.35\linewidth}{@{}lr@{}}
\toprule
  Input & Dimensionality \\
\midrule
  Joint Positions & 16  \\
  Kp Gains  & 16  \\
  Kd Gains & 16  \\
\bottomrule
\end{tabular*}
\caption{\textbf{Action space.} We experimentally validate our approach on a Digit robot which has 20 actuated joints in total. For consistency with the pre-training dataset, we treat the robot feet as passive and control the remaining 16 motors with our model. The model predicts the desired joint positions as well as the P and D gains for each motor, leading to 48 dimensional actions.}
\label{tab:act_state}
\end{center}
\end{table}

\begin{table}
\begin{center}
\begin{tabular*}{0.39\linewidth}{@{}lr@{}}
\toprule
  Parameter & Value \\
\midrule
  Number of GPUs & 4 A100s \\
  Training Epochs & 300 \\
  Minibatch Size &  4096 \\
  Optimizer & AdamW \\
  Learning Rate & 5e-4 \\
  Learning Rate Schedule & cosine \\
  Warm-up Epochs & 30 \\
  Warm-up Schedule & linear \\
  Weight Decay & 0.01 \\
  Optimizer Momentum $\beta_1$ & 0.9 \\
  Optimizer Momentum $\beta_2$ & 0.95 \\
\bottomrule
\end{tabular*}
\caption{\textbf{Pre-training hyperparameters.}}
\label{tab:pre_hyperparams}
\end{center}
\end{table}

\begin{table}
\begin{center}
\begin{tabular*}{0.56\linewidth}{@{}lr@{}}
\toprule
  Parameter & Value \\
\midrule
  Number of GPUs & 1 A10 \\
  Number of Environments & 2048 \\
  Learning Epochs & 5 \\
  Steps per Environment & 24 \\
  Minibatch Size &  12288 \\
  Episode Length & 20 seconds \\
  Discount Factor ($\gamma$) & 0.99 \\ 
  Generalised Advantage Estimation ($\lambda$) & 0.95 \\
  Initial Noise Standard Deviation & 0.135 \\
  PPO Clipping Parameter & 0.2 \\
  Optimizer & AdamW \\
  Learning Rate (Actor) & 1e-5 \\
  Learning Rate (Critic) & 5e-4 \\
  Learning Rate Schedule (Actor) & cosine \\
  Learning Rate Schedule (Critic) & constant \\
  Weight Decay & 0.01 \\
  Training Iterations & 2000 \\
\bottomrule
\end{tabular*}
\caption{\textbf{Fine-tuning hyperparameters.}}
\label{tab:ppo_hyperparams}
\end{center}
\end{table}

\begin{table}
\begin{center}
\begin{tabular*}{0.81\linewidth}{@{}lcccc@{}}
\toprule
  Terrain Type & Fraction (\%) & Unit & Range & Image  \\
\midrule
  Flat  & 12.5  & \% grade & [0, 0] & \includegraphics[width=0.2\linewidth]{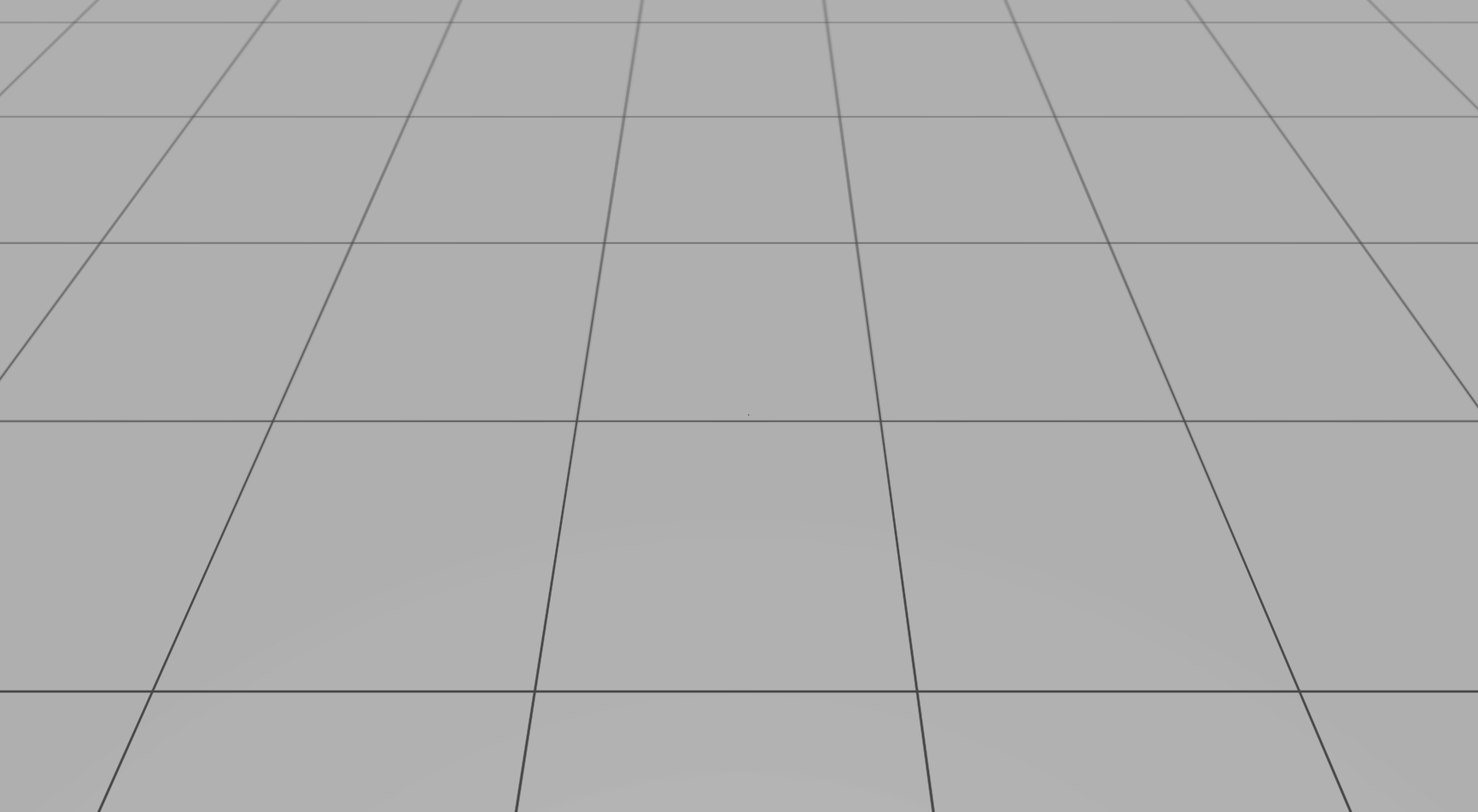} \\
  Rough & 12.5  & cm & [-2.5, 2.5] & \includegraphics[width=0.2\linewidth]{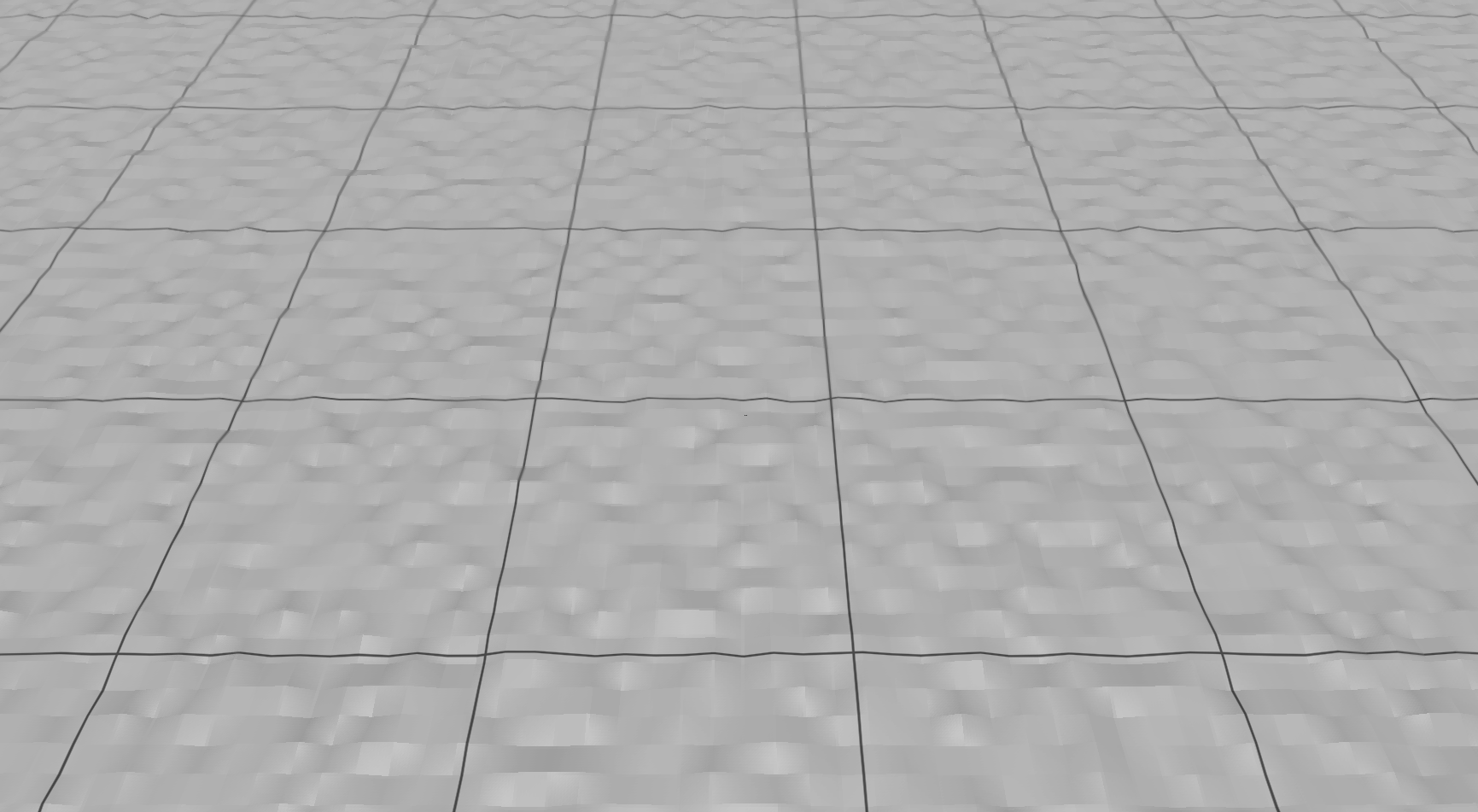}  \\
  Smooth Slope & 12.5 & \% grade & [2, 20] & \includegraphics[width=0.2\linewidth]{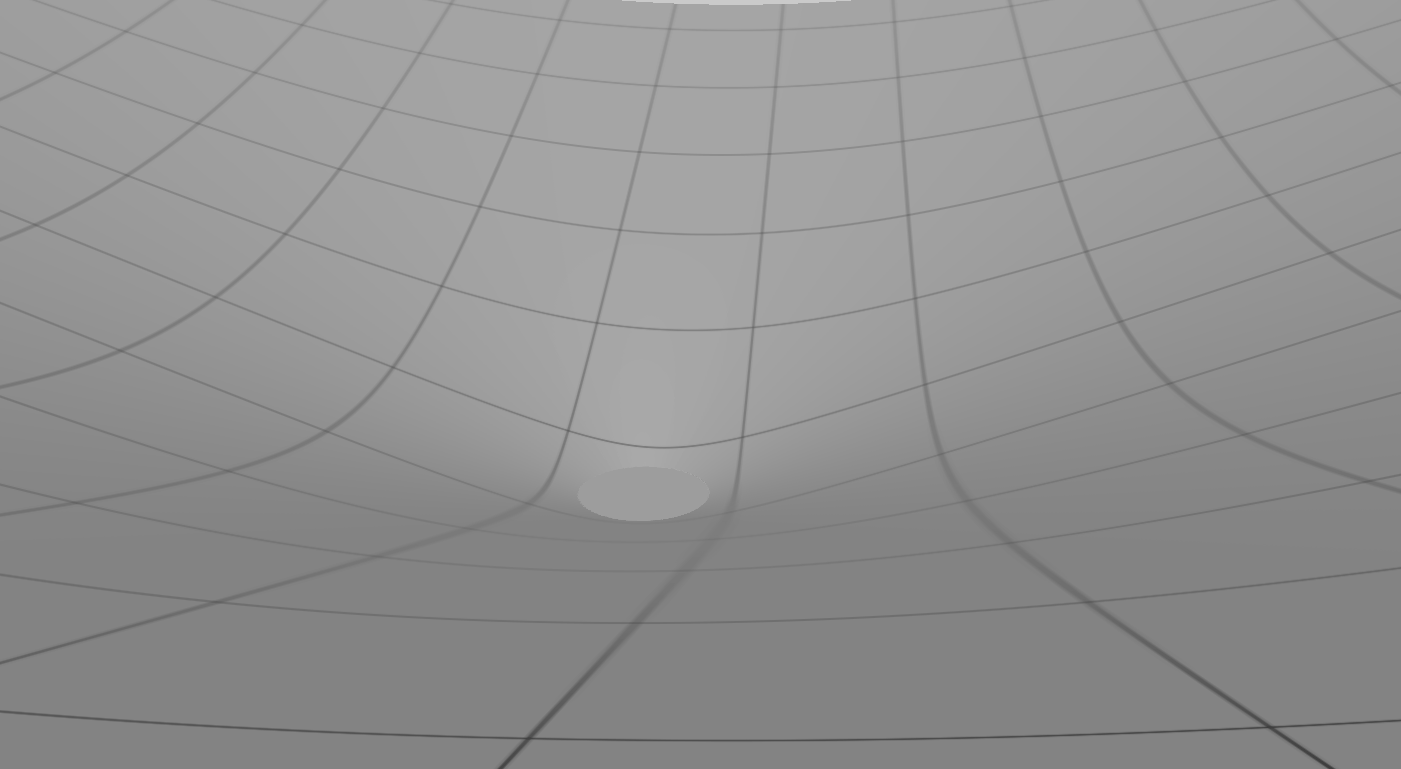} \\
  Rough Slope & 12.5 & \% grade  & [2, 20] & \includegraphics[width=0.2\linewidth]{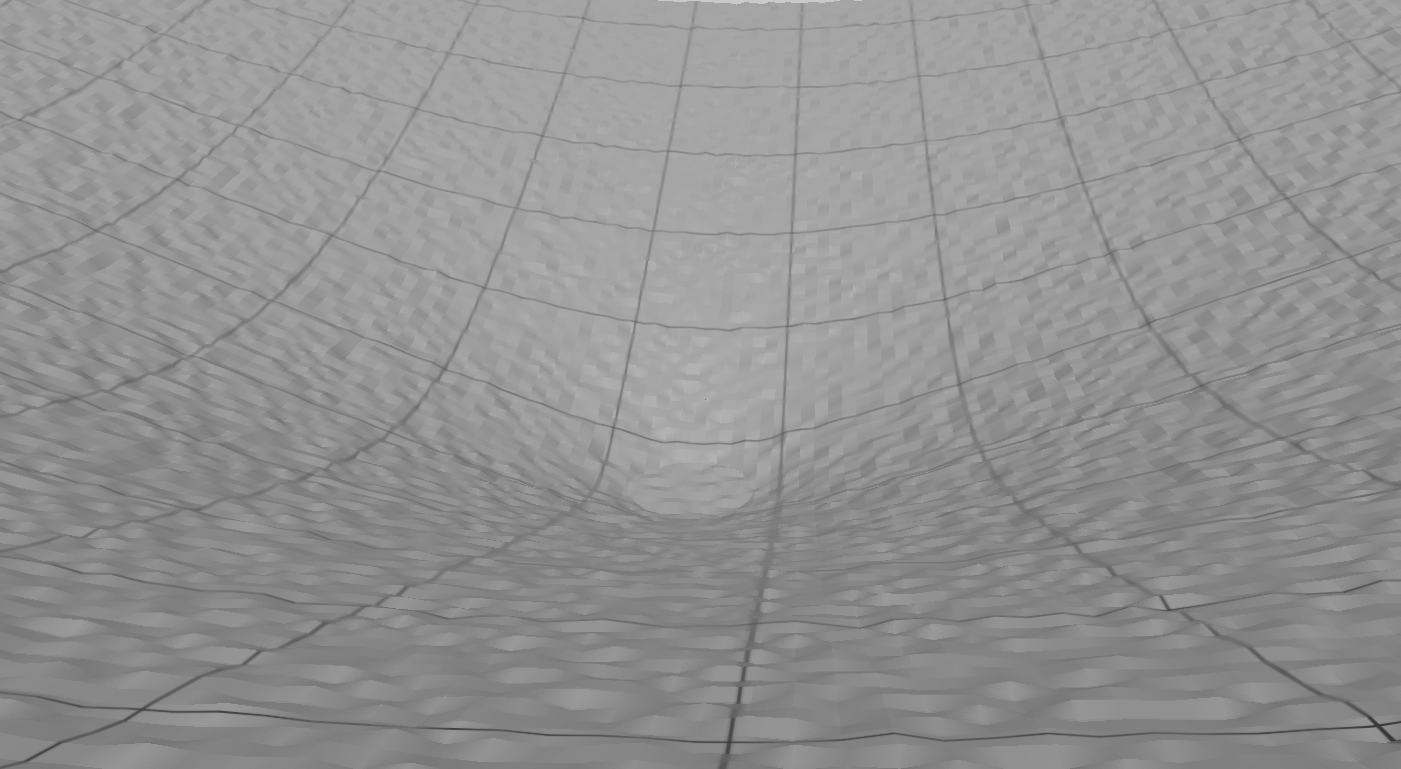} \\
  Obstacles & 25 & cm & [0.5, 5] & \includegraphics[width=0.2 \linewidth]{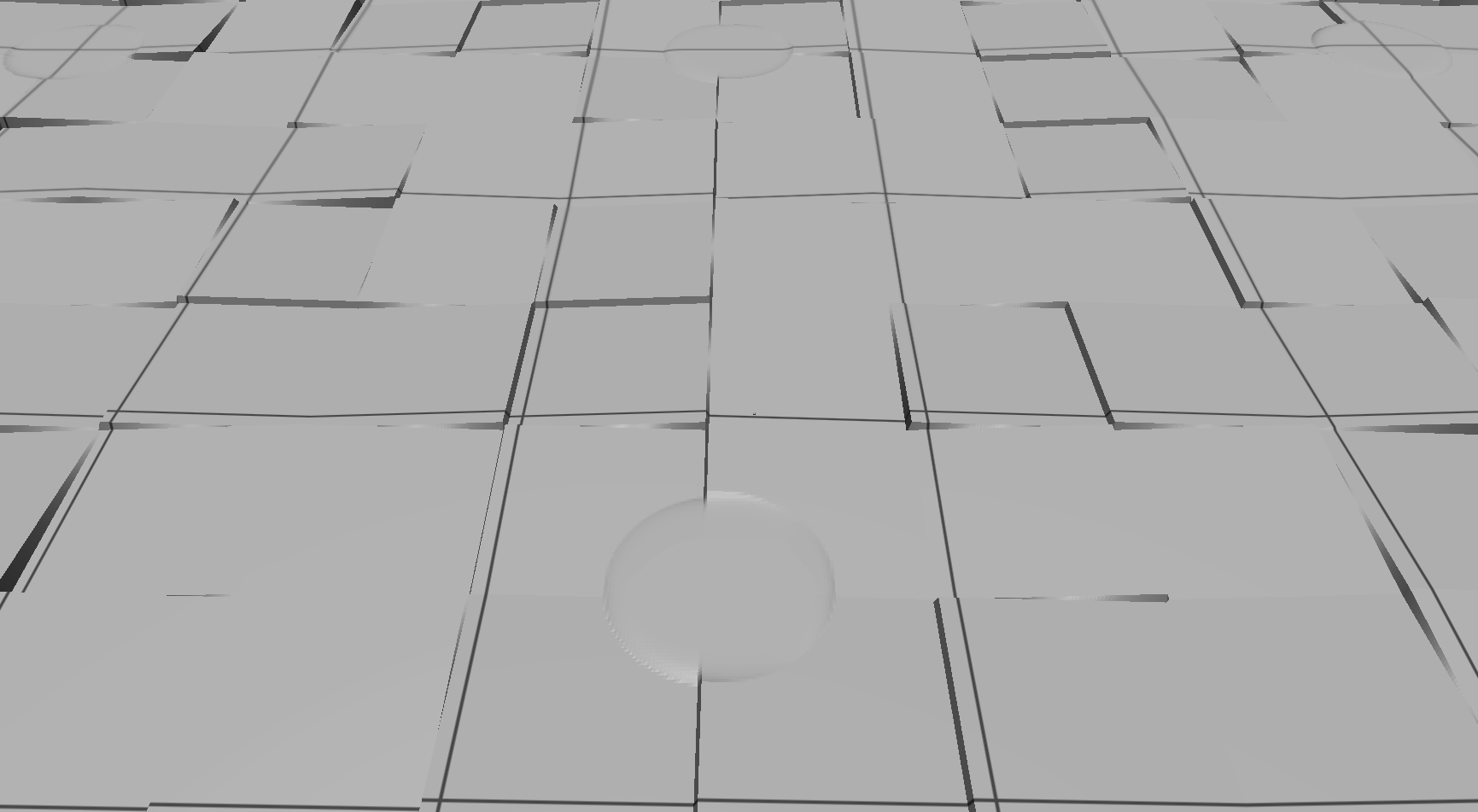} \\
  Hills & 25 & scale & [0.55, 0.75] & \includegraphics[width=0.2 \linewidth]{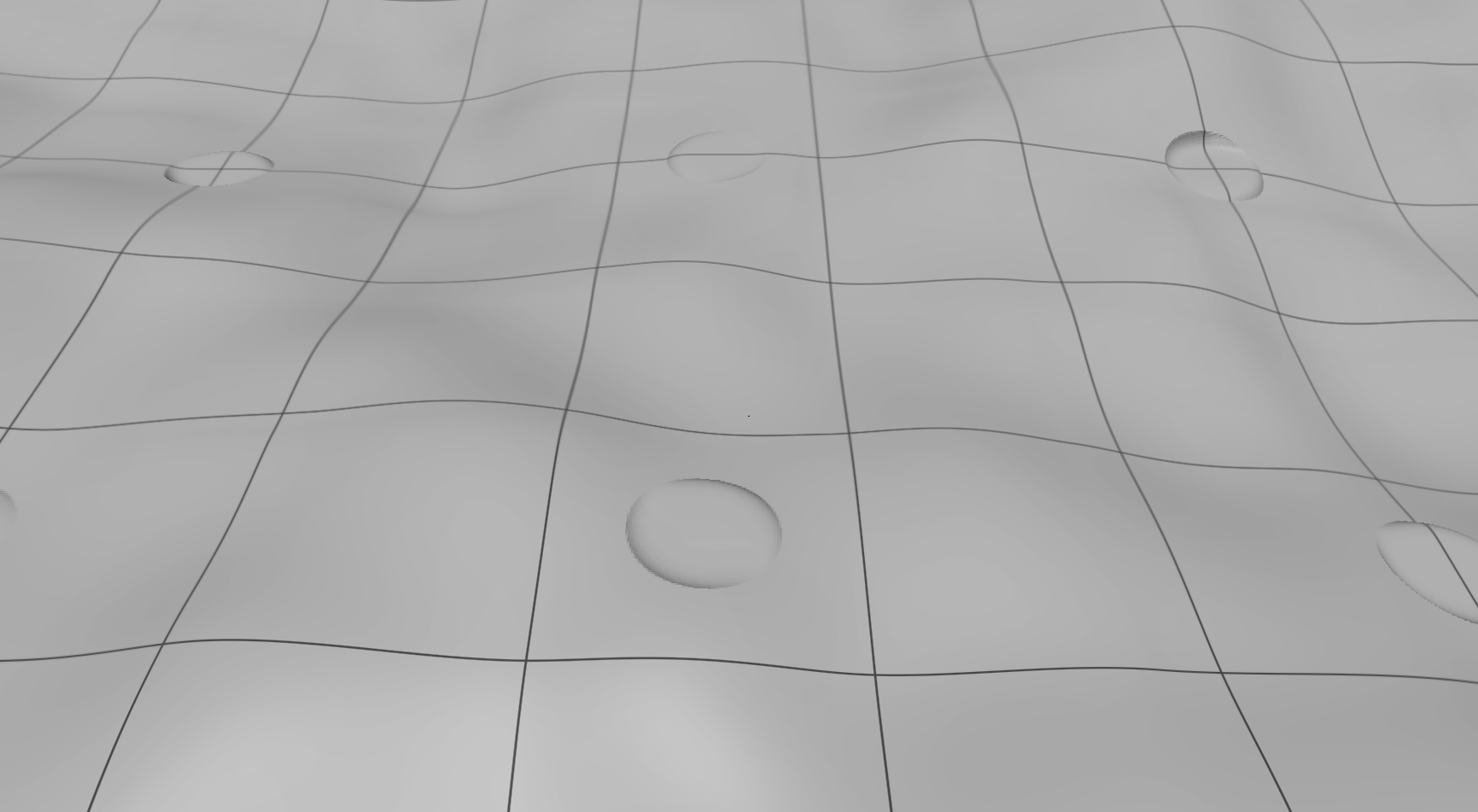} \\
\bottomrule
\end{tabular*}
\caption{
\textbf{Terrains}. We fine-tune the model with reinforcement learning on an ensemble of different terrains in simulation. We use six different terrain types and resample variations of each terrain periodically. Terrain types, sampling ranges, and example images are shown above.}
\label{tab:terrain}
\end{center}
\end{table}

\newpage

\begin{table}
\begin{center}
\begin{tabular*}{0.6\linewidth}{@{}llcccc@{}}
\toprule
  Parameter & Unit & Range & Prob. & Change Int.\\
\midrule
  Forward Speed & m/s   & [-1.0, 1.0] & 0.5 & 10 sec.\\
  Lateral Speed & m/s   & [-0.5, 0.5] & 0.5 & 10 sec.\\
  Turning Speed & rad/s & [-1.0, 1.0] &  0.5 & 10 sec. \\
\bottomrule
\end{tabular*}
\caption{\textbf{Velocity commands}. We sample velocity commands from the provided ranges using a uniform distribution. To make the commands more realistic, we set each dimension to zero with a probability of one-half. The commands are re-sampled periodically, at change intervals.}
\label{tab:command}
\end{center}
\end{table}

\begin{table}
\begin{center}
\begin{tabular*}{0.74\linewidth}{@{}llccc@{}}
\toprule
  Parameter & Unit & Range & Operator & Distribution \\
\midrule
 Joint Position & rad & [0.0, 0.175] & additive & gaussian \\
 Joint Velocity & rad/s & [0.0, 0.15] & additive & gaussian \\
 Base Lin. Vel. & m/s & [0.0, 0.15] & additive & gaussian \\
 Base Ang. Vel. & rad/s & [0.0, 0.15] & additive & gaussian \\
 Gravity Projection & - & [0.0, 0.075] & additive & gaussian \\
 Observation Delay & B(p)$\times$dt & [0.0, 0.2] &   & uniform \\
 Action Delay & B(p)$\times$dt & [0.0, 0.2] &   & uniform \\
 \midrule
 Motor Offset & rad & [0.0, 0.035] & additive & uniform \\
 Motor Strength & \% & [0.85, 1.15] & scaling & uniform \\
 Joint Damping & \% & [0.3, 4.0] & scaling & loguniform \\
 Joint Stiffness & \%  & [0.3, 1.5] & scaling & loguniform \\
 Mass & \% & [0.5, 1.5] & scaling & uniform
 \\
  Size & \% & [0.95, 1.05] & scaling & uniform
 \\
 \midrule
 Kp Factor & \% & [0.9, 1.1] & scaling & uniform \\
 Kd Factor & \% & [0.9, 1.1] & scaling & uniform \\
 \midrule
 Gravity & m/s$^2$ & [0.0, 0.67] & additive & uniform \\
 Friction & \% & [0.3, 2.0] & scaling & uniform \\
 Time Constant & \% & [0.2, 5] & scaling & uniform \\
 Damping Ratio & \% & [0.3, 2] & scaling & uniform \\
\bottomrule
\end{tabular*}
\caption{\textbf{Domain randomization.} Our domain randomization settings largely follow~\cite{RealHumanoid2023}. We sample values from the provided ranges for each parameter. Additive and scaling operators add to and scale the default parameter values, respectively. The ranges specify the [lower, upper] bound and [mean, standard deviation] for the uniform and Gaussian distributions, respectively.}
\label{tab:dr}
\end{center}
\end{table}

\end{appendices}

\end{document}